\documentclass[10pt,twocolumn,letterpaper]{article}

\usepackage{cvpr}              

\definecolor{cvprblue}{rgb}{0.21,0.49,0.74}
\usepackage[pagebackref,breaklinks,colorlinks,allcolors=cvprblue]{hyperref}
\usepackage{multirow}
\usepackage{pifont}
\usepackage{amssymb}
\definecolor{darkgreen}{rgb}{0,0.8,0}
\usepackage{tikz}
\newcommand{\boldcirc}{\tikz \draw[darkgreen, line width=0.5mm] (0,0) circle (0.5ex);}
\usepackage{caption}
\usepackage{MnSymbol}
\usepackage{colortbl}
\usepackage[breakable]{tcolorbox}

\definecolor{mygray}{gray}{0.9}
\definecolor{myyellow}{RGB}{255, 255, 150}

\def\paperID{15800} 
\def\confName{CVPR}
\def\confYear{2025}

\title{CoSpace: Benchmarking Continuous Space Perception Ability\\ for Vision-Language Models}

    \author{\textbf{Yiqi Zhu\textsuperscript{1*}, Ziyue Wang\textsuperscript{1*}, Can Zhang\textsuperscript{3}, Peng Li\textsuperscript{2,4†}, Yang Liu\textsuperscript{1,2,4,5†}} \\
\textsuperscript{1}Dept. of Comp. Sci. \& Tech., Institute for AI, Tsinghua University, Beijing, China \\
\textsuperscript{2}Institute for AI Industry Research (AIR), Tsinghua University, Beijing, China \\
\textsuperscript{3}School of Computer and Communication Engineering, University of Science and Technology Beijing \\
\textsuperscript{4}Shanghai Artificial Intelligence Laboratory, Shanghai, China \\
\textsuperscript{5}Jiangsu Collaborative Innovation Center for Language Competence, Jiangsu, China \\ \\
}

\begin{document}
\twocolumn[{
\maketitle
\vspace{-35pt}
\begin{center}
    \centering
    \captionsetup{type=figure}
    \vspace{-12pt}
    \includegraphics[width=1.0\textwidth]{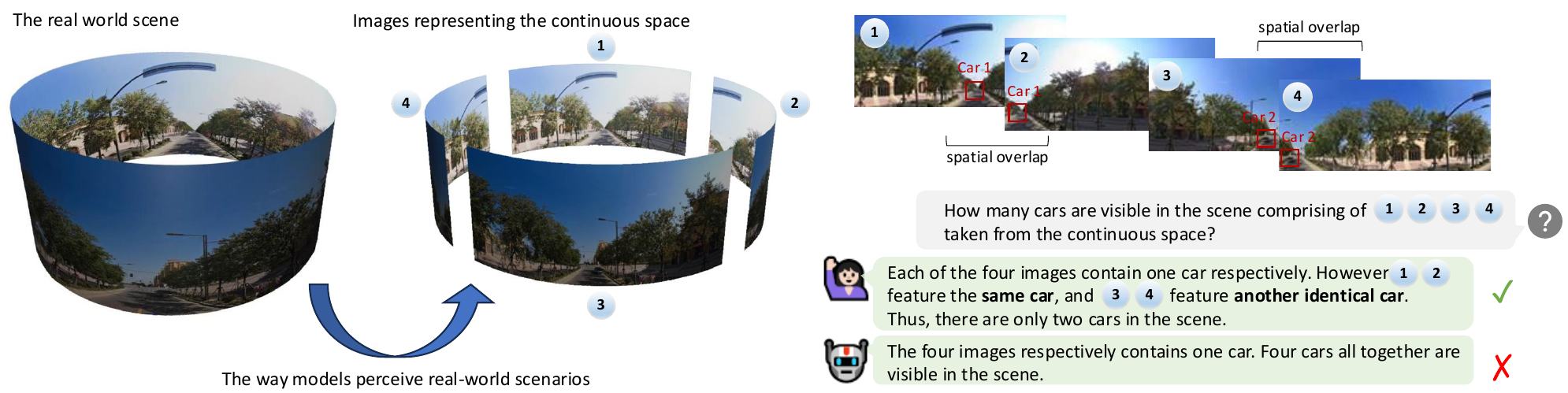}
  \vspace{-18pt}
  \caption{Illustration of continuous space perception: The left side depicts the construction of an image sequence representing a continuous space. The right side presents an example task related to the continuous space, showing how adjacent images are connected. Note that to answer the question correctly, one must recognize that the cars in images 1 and 2 are the same, as well as those in images 3 and 4.}
  \label{fig:teaser}
  \vspace{-3pt}
\end{center}
}]

\newcommand\blfootnote[1]{%
  \begingroup
  \renewcommand\thefootnote{}\footnote{#1}%
  \addtocounter{footnote}{-1}%
  \endgroup
}
\blfootnote{*Equal Contribution, †Corresponding Author}
\blfootnote{\phantom{*}Project Page: \hyperlink{https://thunlp-mt.github.io/CoSpace/}{https://thunlp-mt.github.io/CoSpace/}}
\blfootnote{\phantom{*}Github Page: \hyperlink{https://github.com/THUNLP-MT/CoSpace/}{https://github.com/THUNLP-MT/CoSpace/}}

\vspace{-6pt}
\begin{abstract}
Vision-Language Models (VLMs) have recently witnessed significant progress in visual comprehension. As the permitting length of image context grows, VLMs can now comprehend a broader range of views and spaces. Current benchmarks provide insightful analysis of VLMs in tasks involving complex visual instructions following, multi-image understanding and spatial reasoning. However, they usually focus on spatially irrelevant images or discrete images captured from varied viewpoints. The compositional characteristic of images captured from a static viewpoint remains underestimated. We term this characteristic as \textbf{Continuous Space Perception}. 
When observing a scene from a static viewpoint while shifting orientations, it produces a series of spatially continuous images, enabling the reconstruction of the entire space. In this paper, we present CoSpace, a multi-image visual understanding benchmark designed to assess the \textbf{Co}ntinuous \textbf{Space} perception ability for VLMs. CoSpace contains 2,918 images and 1,626 question-answer pairs, covering seven types of tasks. We conduct evaluation across 19 proprietary and open-source VLMs. Results reveal that there exist pitfalls on the continuous space perception ability for most of the evaluated models, including proprietary ones. Interestingly, we find that the main discrepancy between open-source and proprietary models lies not in accuracy but in the consistency of responses. We believe that enhancing the ability of continuous space perception is essential for VLMs to perform effectively in real-world tasks and encourage further research to advance this capability.

\end{abstract}
\vspace{-2em}
\section{Introduction}
\label{sec:introduction}

With the advancements in deep learning and Large Language Models (LLMs), Vision-Language Models (VLMs) have demonstrated exceptional performance across a range of downstream tasks~\cite{li2023otter, zhang2024long, lin2023sphinx, jia2024leopard, shiri2024empirical, Chen2024CVPR}. Driven by this, VLMs are now being used in diverse applications, such as planning~\cite{chen2023towards,yang2024guidinglonghorizontaskmotion}, navigation~\citep{chiang2024mobility, zhou2025navgpt} and embodied tasks~\citep{Li_2024_CVPR,rajvanshi2024saynav}, and are further applied in real-world scenarios~\citep{zhang2024mmerealworldmultimodalllmchallenge,cao2024maplm}. Real-world scenario, featuring the complicated environment, cast significant challenges on VLMs, especially the ability to capture sophisticated spatial information from surrounding spaces.

\begin{table*}[!t]
\centering
\scriptsize
\begin{tabular}{l|l|@{\hspace{0.1cm}}c@{\hspace{0.25cm}}c@{\hspace{0.25cm}}c@{\hspace{0.1cm}}|cc@{\hspace{0.1cm}}}
\toprule
\multirow{3}{*}[-1ex]{Benchmarks} & \multirow{3}{*}[-1ex]{Evaluation Target} & \multicolumn{3}{c|}{Features} & \multirow{3}{*}[-1ex]{\begin{tabular}[c]{@{}c@{}}Num.\\Img\end{tabular}} & \multirow{3}{*}[-1ex]{\begin{tabular}[c]{@{}c@{}}Evaluation\\Instances\end{tabular}} \\ \cmidrule(lr){3-5}
 &  & {\begin{tabular}[c]{@{}c@{}}Spatial\\Reasoning\end{tabular}}& {\begin{tabular}[c]{@{}c@{}}Multi-image\\Perception\end{tabular}} & {\begin{tabular}[c]{@{}c@{}}Static\\Viewpoint\end{tabular}} &  \\ 
 \midrule
 MMRel~\citep{nie2024mmrel}& Object-relationship & \boldcirc & \textcolor{red}{\ding{55}} & \textcolor{red}{\ding{55}} & 15k & 15k \\ 
 Spatial-MM~\citep{shiri2024empirical} & Object-relationship and spatial-CoT & \textcolor{darkgreen}{\ding{51}} & \textcolor{red}{\ding{55}} & \textcolor{red}{\ding{55}} & 2.3k & 400\\
 SpatialEval~\citep{wang2024is} & Spatial Reasoning & \textcolor{darkgreen}{\ding{51}} &\textcolor{red}{\ding{55}} & \textcolor{red}{\ding{55}} & 4.6k & 4.6k\\
 GSR-Bench~\citep{rajabi2024gsr} & Grounded spatial reasoning & \textcolor{darkgreen}{\ding{51}} & \textcolor{red}{\ding{55}} & \textcolor{red}{\ding{55}} & 4.9k & 4.9k \\ 
 MileBench~\citep{dingjie2024milebench}& Multimodal Long Context & \textcolor{red}{\ding{55}} & \textcolor{darkgreen}{\ding{51}} & \textcolor{red}{\ding{55}} & 97.9k & 6.4k\\
 Demon-Core~\citep{li2024finetuning}& Demonstrative instruction following & \textcolor{red}{\ding{55}} & \textcolor{darkgreen}{\ding{51}} & \textcolor{red}{\ding{55}}  & 62.8k & 18.2k\\
 Mementos-Val~\citep{wang-etal-2024-mementos}& Temporal sequential-image understanding & \textcolor{red}{\ding{55}} & \textcolor{darkgreen}{\ding{51}} & \textcolor{red}{\ding{55}} & - & 699\\
 MIBench~\citep{liu2024mibench} & Fine-grained multi-image understanding & \textcolor{red}{\ding{55}} & \textcolor{darkgreen}{\ding{51}} & \textcolor{red}{\ding{55}} & 125k & 13k\\
 BLINK~\citep{fu2024blink} & General visual perception & \boldcirc & \boldcirc & \textcolor{red}{\ding{55}} & 7.3k & 3.8k\\ 
 SEED-Bench~\citep{li2024seed} & Visual comprehension  & \boldcirc & \boldcirc & \textcolor{red}{\ding{55}} & 1.9k & 24k\\ 
 Video-MME~\citep{fu2024video} & Visual comprehension & \boldcirc & \textcolor{darkgreen}{\ding{51}} & \textcolor{red}{\ding{55}} & 0.9k & 2.7k \\ 
 MMIU~\citep{meng2024mmiu} & Visual comprehension & \boldcirc & \textcolor{darkgreen}{\ding{51}} & \textcolor{red}{\ding{55}} & 77.7k & 11.7k \\ 
 SparklesEval~\citep{huang2024sparkles} & Multi-image Chat & \boldcirc & \textcolor{darkgreen}{\ding{51}} & \textcolor{red}{\ding{55}} & 550 & 150 \\   
\midrule
CoSpace (Ours) & continuous space perception & \textcolor{darkgreen}{\ding{51}} & \textcolor{darkgreen}{\ding{51}} & \textcolor{darkgreen}{\ding{51}} & 2,918 & 1,626 \\
\bottomrule
\end{tabular}
\vspace{-1em}
\caption{Comparison with selected benchmarks that address multi-image evaluation and/or spatial evaluation. {\textcolor{darkgreen}{\ding{51}}} denotes the main focus of the corresponding feature, \protect\boldcirc ~indicates that the feature exists but is not the primary focus, while {\textcolor{red}{\ding{55}}} implies a lack of the feature.}
\label{tab:compare}
\vspace{-2em}
\end{table*}

Despite previous attempts to advance spatial understanding in VLMs~\citep{Chen2024CVPR, nie2024mmrel}, 
there remain areas requiring further exploration. Various strategies have been explored to empower models with the ability to identify and ground objects based on spatial relationship~\cite{Chen2024CVPR, jia2024sceneverse}. \citet{shiri2024empirical} and \citet{wang2024is} assess spatial understanding ability for VLMs with the recognition of basic tasks in this domain. However, much of these work center on single-image inputs, without further extending to multi-image scenarios. Works on video understanding~\citep{zhang2024long,wang2024longllava} present the concept of spatial temporal understanding, but the spatial information of videos involve sequential frames that inherently linking spatial understanding with additional temporal dynamics. Given these attempts in spatial understanding, we pose the question: {\it Are the current efforts adequate for advancing the research of spatial understanding?} 

\vspace{-3pt}
We claim that a gap still exists between the concept proposed in previous work and more advanced spatial understanding. For humans, exposing to an unfamiliar environment requires determining the position, identifying surrounding objects, and making decisions based on spatial context. 
Such process typically involves integrating various information about the surrounding space. We refer to it as \textbf{continuous space perception},
which is an essential ability in real-world towards effectively completion of daily tasks. 

\vspace{-3pt}
Figure~\ref{fig:teaser} shows an example of continuous space perception: Conceptually, humans perceive continuous visual space as a series of ``images'', each representing a segment of the space with overlaps between adjacent views. Mimicking this perception, Figure~\ref{fig:teaser} illustrates the continuous visual space using four images. To answer questions accurately about these images requires a comprehensive understanding of all four images and reconstructing the entire space they represent. For example, to accurately count the number of cars in the space, it is essential to recognize that the cars appearing in the first and second images are the same, as are the cars in the third and fourth images.

\vspace{-3pt}
The continuous space perception ability is equally important for VLMs as for human. Many real-world scenario tasks, such as planning and navigation, require models to effectively integrate the information in the surroundings to take the optimal actions. During this process, the reconstruction of the space heavily relies on continuous space perception. However, as shown in Table~\ref{tab:compare}, this ability is largely overlooked in existing benchmarks for VLMs.

To this end, we present \textbf{CoSpace}, a visual understanding benchmark designed to evaluate the \textbf{Co}ntinuous \textbf{Space} perception ability for VLMs. CoSpace includes four categories and seven tasks, emphasizing the following core abilities: (1) Can VLMs recognize direction and position when presented with spatially continuous images captured from a static viewpoint? (2) Can models identify the correspondence between adjacent images in the same scene? (3) Can models make optimal navigation decisions based on the perception of the space? We posit the importance of these three core abilities, with the first one forms the basic understanding of space, the second one is essential for integrating the information across different images and the last one surpasses mere literal understanding and makes a step towards more practical applications.

We evaluated 19 widely used open-source and proprietary VLMs on our proposed CoSpace benchmark. Results show that though some of the open-source models achieve approaching or even higher accuracy compared with state-of-the-art proprietary models on certain tasks, proprietary models exhibit remarkably higher consistency in their responses across multi-dimensional evaluation metrics. Additionally, we found that with the alteration of input context and image sequence, the tasks that humans find more challenging do not always pose greater difficulty for the models.

\vspace{-3pt}
\section{Related Work}
\label{sec:relatedwork}

\subsection{Multi-image Perception Ability of VLMs}
Large Vision-Language Models (VLMs), such as LLaVA models~\citep{liu2023llava, liu2024llavanext} and Sphinx~\citep{lin2023sphinx}, have shown impressive performances on visual reasoning and grounding tasks that focus on only one image at a time. However, it is more natural to allow multiple images and arrange them in interleave format in one input. Along with the growing capabilities of single-image models, multi-image perception and understanding abilities are gaining increasing attentions, including research topics such as multimodal in-context learning~\citep{li2023otter, zhao2024mmicl, doveh2024towards}, interleaved visual instruction following~\citep{jiang2024mantis, li2024finetuning, li2024llavanextinterleave}, and temporal visual understanding~\citep{tang2023video}, and practical applications such as navigation and embodied QA~\citep{zhou2025navgpt}. 
To achieve multi-image perception, some methods focus on constructing interleaved visual instruction tuning data~\citep{li2023otter, zhao2024mmicl, jiang2024mantis, jia2024leopard}, some improve the model structure and training strategies for better cross-image understanding~\citep{li2024finetuning, wang-etal-2024-browse, ye2024mplug3, wang2024longllava}, and others address the image processing strategy for adaptable visual feature encoding concerning multi-image scenarios~\citep{liu2024llavanext, laurençon2024idefics3, zhang2024mm15methodsanalysis}. 

With the enhancements of the multi-image perception ability, VLMs are able to obtain visual information covering a wider range of views from multiple images at one time, resulting in more profound observations and understandings of the surrounding environment. In our paper, we take this as basis, and dedicate to investigating the spatial perception ability of VLMs.

\subsection{Spatial Comprehension Evaluation of VLMs}
For the past years, the evaluation of spatial comprehension for VLMs has always been a commonly engaged topic, and it often refers to the spatial relation recognition among objects, such as MMRel, GSR-Bench and Spatial-MM shown in Table~\ref{tab:compare}. 
Recently, multi-image related evaluation for VLMs are widely explored, including interleaved visual instruction following~\citep{li2024finetuning, liu2024mibench}, temporal sequential-image comprehension~\citep{wang-etal-2024-mementos}, and long context understanding~\citep{dingjie2024milebench}. In multi-image scenarios, such as video comprehension~\citep{fu2024video, li2024seed} and multi-image dialogue~\citep{huang2024sparkles}, the spatial reasoning ability is also one of the evaluated aspects. However, these benchmarks primarily focus on general multi-image comprehension where most of input images come from difference scenes. The compositional features across multiple images to integrate richer spatial information of a wider scene and more profound spatial understanding are seldom emphasized. 
Although Blink~\citep{fu2024blink}, ActiView~\citep{wang2024actiview}, and video-based benchmarks~\citep{fu2024video,li2024mvbench} require spatial understanding across multiple images, they either focus on different views of the same object, or discrete views from dynamic viewpoint. These views are not continuous regarding a static viewpoint, and the resulting images are not necessarily bounded regarding the same scene.
In contrast to existing works, we address the continuous spatial perception from a static viewpoint, that requires models to connect all the perceived views together to form an intact understanding of the whole scene.

\begin{figure*}
    \centering
    \includegraphics[width=1.\linewidth]{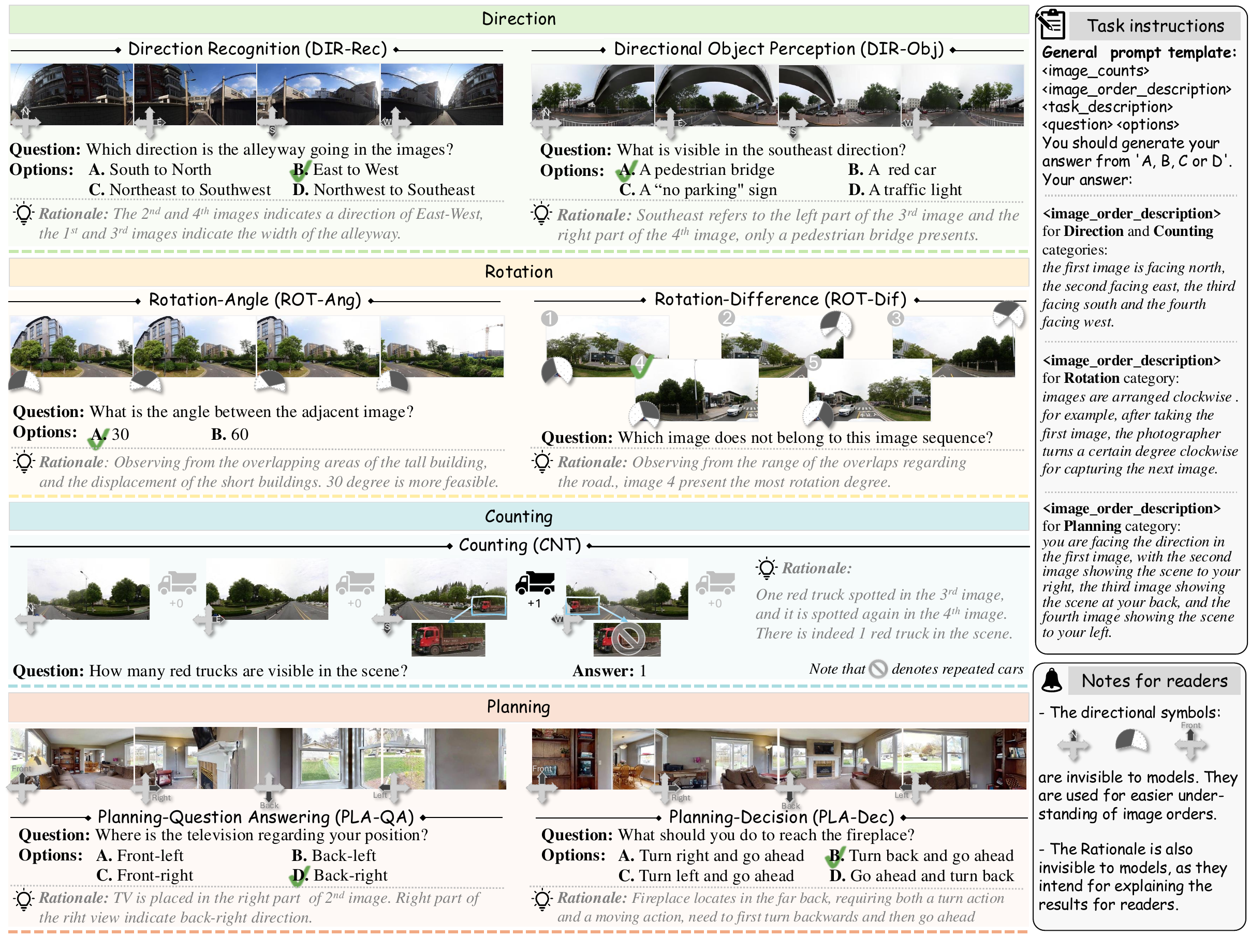}
    \vspace{-2.5em}
    \caption{Examples of CoSpace, along with prompt templates for evaluation. We also provide extra visual guidance and textual rationales for readers (invisible to models during evaluation) for easier understanding.}
    \label{fig:cases}
    \vspace{-1em}
\end{figure*}

\section{CoSpace}
\label{sec:methodology}

\subsection{Continuous Space Perception Ability}
\label{sec:spaceperception}

In this paper, we define the term, {\bf continuous visual space}, as the perceived views from a static viewpoint towards different orientations that can naturally compose a complete scene or environment. As is shown in Figure~\ref{fig:teaser}, if one would like to correctly answer the question, it is needed to first capture the visual information of all the cars, and 
consequently, the ability of {\bf continuous space perception} refers to capturing and processing information from these images of continuous visual spaces. 

To obtain an overall observation of environment, human perceive the whole surroundings from the same viewpoint towards different directions, and the percept continuous visual spaces can be reconstructed as a panorama in the brain.
Different from human that directly views the space continuously, models typically take discrete images as input. Therefore, the way to present continuous visual spaces remains flexible. In our CoSpace, we present a complete continuous visual space as follows: we provide several images shot at the same viewpoint in the same scene, but towards different directions, e.g. respectively towards front, right, back and left. These images together form a complete space without omitting any visual information of the surrounding environment. Naturally, such strategy could produce overlapped spaces between adjacent images, which tightly connects images describing adjacent spaces, and can consequently help models better identify the relationships between images from continuous visual space. In this way, we posit that such strategy can effectively preserve continuous spatial information of an extensive space as well as be suitable to understanding for VLMs.

\subsection{Task Design}
\label{sec:CoSpace}

We comprehensively evaluate the continuous space perception ability of VLMs regarding four aspects. Two of them are fundamental ability of spatial perception, detailed as direction perception and recognition (\textit{Direction}), and rotation-centric perception (\textit{Rotation}). The other two are further practical tasks, detailed as counting-based perception (\textit{Counting}), and embodied-based planning (\textit{Planning}). These four categories are subdivided into seven tasks for elaborate investigation as shown in Figure~\ref{fig:cases}.

We employ two multiple choice tasks, both with four options, for the \textit{Direction} category as follows:
\begin{itemize}
    \item \textbf{Direction Recognition (DIR-Rec)}: In real world, identifying direction is inevitable when one is placed in a new environment. Similarly, the DIR-Rec task requires models to recognize the direction of targets by answering questions such as ``\textit{where is the building located?}'' Meanwhile, we provide detailed task instructions, including direction references such as ``\textit{the first image is facing north, the second image is facing east}''. We standardize 8 directional options, including four cardinal directions, North, East, South and West, and four intercardinal directions, Northeast, Southeast, Southwest, and Northwest.
    \item \textbf{Directional Object Perception (DIR-Obj)}: Some real-world tasks demand abilities far beyond merely identifying the direction of a specified target. It is also a critical ability to distinguish content regarding different directions and to align objects with correct directions. Given a specified direction, the DIR-Obj task requires models to be aware of the appearing objects, and also distinguish those absent from that direction. To achieve this, models need to first reconstruct the entire space from continuous views, and then correctly identify the objects dedicate to a given direction. Specifically, in this task, questions typically focus on the intercardinal directions such as ``\textit{What is visible in the southeast direction?}'', which necessitates a more comprehensive understanding of the entire space.
\end{itemize}

Observing the surroundings from a static viewpoint naturally involves rotating the perspectives. Therefore, we summaries \textit{Rotation} as another fundamental category of continuous space perception, with two tasks designed as follows:
\begin{itemize}
    \item \textbf{Rotation-Angle (ROT-Ang)}: For an observant, when rotating the perspective for more spatial information, there could be overlaps between adjacent observations. Specifically, human can tell the approximate turning angle between consecutive images observed from a static viewpoint by noticing identical furnishings appearing across images. In this task, we investigate if models exhibit similar ability as human does, which requires fine-grained continuous spatial understanding. We apply unified question ``\textit{What is the turning angle between the adjacent image?}'' and provide two candidate options for model. This task requires models to carefully exam the overlapping and different areas between adjacent images and the shared features in spatially continuous image sequence.
    \item \textbf{Rotation-Difference (ROT-Dif)}: Following the ROT-Ang task, we also investigate the model ability to identify distinguished rotation angles. For the ROT-Dif task, models are given a sequence of five images, four of which share the same turning angle and the other is exceptional. Compared to ROT-Ang, this task emphasizes more on the global understanding of images regarding continuous visual space. The unified question of this task is ``\textit{Which image does not belong to this image sequence?}'' It is an open-ended question without options, where models are required to answer with index of the exceptional image. 
\end{itemize}

On the basis of the above two fundamental categories, we further challenge models with two practical categories. The \textit{Counting} category addresses the occurrence of objects in the entire scene. The \textit{Planning} category comprises two spatial-related tasks widely evaluated in embodied AI tasks. For the Counting category, the task is described as:

\begin{itemize}
    \item \textbf{Counting (CNT)}: Counting is a widely adopted task in existing visual benchmarks that asks models to recognize objects and deduce the time of occurrence of target objects. Generally, existing counting tasks only require models to deal with a single image or multiple spatially discrete images, while in our benchmark, images from continuous visual space where identical objects can occur in multiple images are focused on. This raise a challenge for models to not only recognize targets and count for their occurrences, but also be aware of the existence of the same object across different images. To achieve this, models should locate the overlapping area of adjacent images and align the same object appearing different images. The CNT task is an open-ended task, where models should response with the total count of the target.
\end{itemize}

The ability of continuous space perception is also significant in embodied scenarios. Intuitively, regarding the location of an embodied agent as the static viewpoint, it should have a profound understanding of the overall surroundings to effectively solve questions, and to efficiently plan for the following action sequence. Embodied question-answering (EQA)~\citep{majumdar2024openeqa, ren2024explore} and navigation~\citep{kim2024realfred, khanna2024goatbench} are typical tasks that require understand continuous visual space. Following these, we split the Planning category into two tasks addressing the relative locations of targets regarding the VLM-based agent as follows:
\begin{itemize}
    \item \textbf{Planning-Question Answering (PLA-QA)}: Following the implementation of EQA, we develop the PLA-QA task, requiring models to identify the location of a certain object given a continuous embodied space. In this task, instructions like ``\textit{Where is the television regarding your position?}'' are provided for the models, and we formulate four options, containing candidate directions relative to the agent, for each question.
    \item \textbf{Planning-Decision (PLA-Dec)}: This task further investigate the understanding of continuous visual space by asking models to select the proper route to reach the target object. For disambiguation, we standardize the action space as turning (turning to other directions without displacement) and go ahead. The PLA-Dec task especially focus on the order of actions. As is shown in the case in Figure~\ref{fig:cases}, the options ``\textit{B. Turn back and go ahead}'' and ``\textit{D. Go ahead and turn back}'' represent two totally different actions and end up in different positions.
\end{itemize}

\begin{table*}[t]
    \centering
    \scriptsize
    \begin{tabular}{@{\hspace{0.1cm}}l@{\hspace{0.1cm}}|c@{\hspace{0.1cm}}c|c@{\hspace{0.1cm}}c|c@{\hspace{0.1cm}}c|c|c|c|c|@{\hspace{0.1cm}}c@{\hspace{0.1cm}}}
        \toprule
        \multirow{2}{*}[-1ex]{Models} & \multicolumn{2}{c|}{DIR-Rec} & \multicolumn{2}{c|}{DIR-Obj} & \multicolumn{2}{c|}{CNT} & \multicolumn{1}{c|}{ROT-Ang} & \multicolumn{1}{c|}{ROT-Dif} & \multicolumn{1}{c|}{PLA-QA} & \multicolumn{1}{c|@{\hspace{0.1cm}}}{PLA-Dec} & \multirow{2}{*}[-1ex]{Average} \\ \cmidrule(lr){2-3} \cmidrule(lr){4-5} \cmidrule(lr){6-7} \cmidrule(lr){8-8} \cmidrule(lr){9-9} \cmidrule(lr){10-10} \cmidrule(lr){11-11}
         & \textit{ACC$_q$} &\textit{ACC$_p$} & \textit{ACC$_q$} & \textit{ACC$_p$} & \textit{ACC$_q$} & \textit{ACC$_p$} & \textit{ACC} & \textit{ACC} & \textit{ACC} & \textit{ACC} \\
         \midrule\noalign{\vskip -3pt}
         \multicolumn{12}{c}{\cellcolor[HTML]{EFEFEF} \scriptsize\textit{Proprietary Models}} \\
         \noalign{\vskip 1pt}
         Claude-3.7-sonnet~\citep{TheC3} & \textbf{44.40} & \underline{29.20} & 45.60 & 35.60 & \underline{45.00} & 38.00 & \textbf{64.33} & \textbf{93.50} & \textbf{54.73} & \textbf{69.34} & \textbf{51.97} \\
         Gemini-1.5-pro~\citep{reid2024gemini} & 37.60 & 15.60 & 40.60 & 31.60 & 38.25 & 24.00 & \underline{59.67} & \underline{82.00} & 51.64 & \underline{62.91} & 44.39 \\
         GPT-4o~\citep{gpt-4o} & \underline{40.40} & 22.80 & 46.00 & 32.00 & 40.00 & 23.50 & 58.33 & 50.50 & \underline{53.05} & 54.46 & 42.10 \\
         \midrule\noalign{\vskip -3pt}
         \multicolumn{12}{c}{\cellcolor[HTML]{EFEFEF}
         \scriptsize\textit{$>$70B Open-source Models}} \\
         \noalign{\vskip 1pt}
         InternVL2\_5-78B~\citep{chen2024expanding} &  32.20 & 24.40 & \textbf{54.40} & \textbf{47.60} & \textbf{51.25} & \underline{43.00} & 50.00 & 77.00 & 32.39 & 42.72 & \underline{45.50} \\
         Qwen2-VL-72B~\citep{wang2024qwen2vl} & 31.00 & 23.20 & \underline{53.60} & \underline{44.80} & 44.75 & 37.00 & 50.00 & 62.00 & 42.72 & 59.15 & 44.82 \\
         InternVL2-76B~\citep{chen2024far} & 33.60 & 13.20 & 46.40 & 39.20 & 50.00 & \textbf{43.50} & 50.00 & 23.00 & 25.35 & 30.05 & 35.43 \\
         LLaVA-OneVision-72B~\citep{li2024llavaov} & 19.20 & 10.80 & 44.20 & 34.40 & 24.25 & 19.00 & 50.00 & 26.00 & 32.86 & 30.99 & 29.17 \\
         \midrule\noalign{\vskip -3pt}
         \multicolumn{12}{c}{\cellcolor[HTML]{EFEFEF} \scriptsize\textit{$<$13B Open-source Models}} \\
         \noalign{\vskip 1pt}
         MiniCPM-V 2.6~\citep{yao2024minicpm} & 32.80 & 21.20 & 40.40 & 31.60 & 38.50 & 31.50 & 50.00 & 56.00 & 41.31 & 27.70 & 37.10 \\
         Qwen2-VL-7B~\citep{wang2024qwen2vl} & 26.40 & 16.40 & 39.20 & 31.60 & \underline{45.00} & 36.00 & 50.00 & 51.50 & 34.27 & 26.76 & 35.71 \\
         Mantis-8B~\citep{jiang2024mantis} & 30.60 & 24.40 & 31.20 & 28.00 & 41.75 & 36.50 & 50.00 & 39.50 & 33.33 & 27.70 & 34.30 \\
         InternVL2-8B~\citep{chen2024far} & 29.00 & 12.80 & 37.20 & 30.40 & 38.50 & 31.00 & 50.33 & 47.50 & 32.39 & 27.70 & 33.68 \\
         VILA1.5-8B~\citep{lin2024vila} & 34.00 & \textbf{30.40} & 28.20 & 24.00 & 42.25 & 38.00 & 50.00 & 18.50 & 22.07 & 42.72 & 33.01 \\
         Idefics3-8B~\citep{laurençon2024idefics3} & 34.00 & 21.20 & 38.60 & 28.00 & 32.75 & 25.00 & 48.00 & 28.00 & 26.76 & 25.35 & 30.77 \\
         LLaVA-OneVision-7B~\citep{li2024llavaov} & 18.80 & 12.00 & 37.60 & 33.60 & 37.75 & 33.00 & 50.00 & 21.50 & 29.58 & 26.29 & 30.01 \\
         Phi-3.5-vision~\citep{abdin2024phi3} & 20.40 & 12.00 & 33.20 & 28.40 & 36.75 & 34.00 & 50.00 & 22.50 & 34.74 & 26.76 & 29.88 \\
         Brote-IM-XXL~\citep{wang-etal-2024-browse} & 33.00 & 13.20 & 31.00 & 24.80 & 30.75 & 29.50 & 50.00 & 18.00 & 26.76 & 20.19 & 27.72 \\
         LongVA-7B~\citep{zhang2024long} & 23.00 & 16.00 & 31.60 & 28.00 & 28.00 & 22.00 & 48.67 & 21.50 & 19.25 & 30.99 & 26.90 \\
         Mono-InternVL-2B~\citep{luo2024monointernvl} & 30.00 & 27.60 & 32.00 & 27.60 & 14.00 & \hphantom{0}8.50 & 50.00 & 18.00 & 25.35 & 24.88 & 25.79 \\
         mPLUG-Owl3-7B~\citep{ye2024mplug3} & 22.80 & \hphantom{0}6.80 & 29.20 & 15.60 & 22.25 & \hphantom{0}9.00 & 46.33 & 16.50 & 27.70 & 21.13 & 21.73 \\
         \bottomrule
    \end{tabular}
    \vspace{-1em}
    \caption{Main results on our benchmark. The proprietary models are accessed via APIs, and the open-source models are accessed from Huggingface checkpoints.}
    \label{tab:main_res}
    \vspace{-1em}
\end{table*}

\begin{figure}
    \centering
    \includegraphics[width=1.\linewidth]{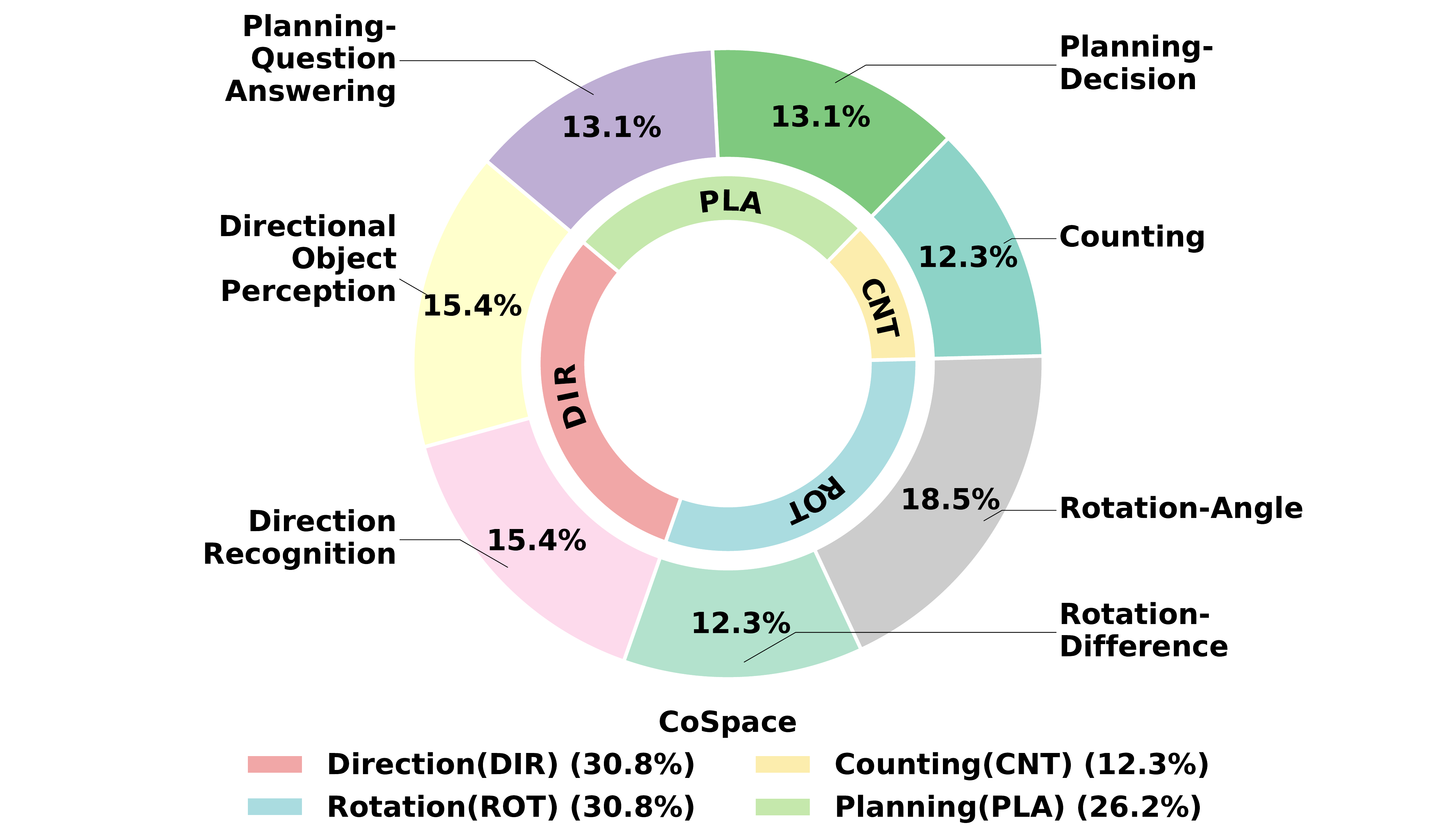}
    \vspace{-2em}
    \caption{Distribution of categories and tasks in our CoSpace.}
    \vspace{-1.2em}
    \label{fig:enter-label}
\end{figure}

\subsection{Data Collection and Statistics}
\label{sec:statistic}

The construction of CoSpace follows the collect-filter-annotate pipeline. We first adopt automatic collection approaches to gather images, and then manually select images that satisfy detailed requirements of our designed tasks and generate questions for these images. Finally human annotators provide answers to each questions. The process is detailed in the following of this section.

Firstly, to automatically collect sets of images captured in the same scene at static viewpoint but towards different directions, we employ two primary sources, Baidu Map Panorama API\footnote{\url{https://lbsyun.baidu.com/}} that allows us to acquire data towards any desired directions in the real world scene, and HM3D~\citep{ramakrishnan2021hm3d}, a dataset of 1,000 building-scale 3D reconstructions for embodied tasks. For HM3D, by using Habitat-Sim~\citep{habitat19iccv, szot2021habitat, puig2023habitat3}, we instruct an agent wandering in the habitats, we are able to collecte images of any views through the attached camera. Secondly, we manually filtered the auto-collected images according to our task requirements, ensuring that the selected images all contain sufficient visual information. With the filtered images, we employed GPT-4o to generate questions via few-shot prompting for the required tasks. Finally, human annotators, an average of two annotators per question, provided answers to these generated questions. This process allowed us to discard vague or unanswerable questions, ensuring that only high-quality question-answer pairs were included in our proposed CoSpace.

Our CoSpace comprises a total of 1,626 question-answer pairs and 2,918 images. Among the images, 2,302 are accessed from Baidu Map API and 616 are derived from HM3D dataset. The distribution of instances over categories and tasks is shown in Figure~\ref{fig:enter-label}. 

\section{Experiments}
\label{sec:evaluation}

As discussed above, CoSpace features reconstructing the surrounding space via multiple images, with each image representing a different space segments. Different from previous multi-image benchmark involving spatial reasoning tasks, the input order of images is crucial in our tasks. To enable that models can associate each image with a specific direction to answer questions in our benchmark, we provide direction references along with images following the prompt templates in Figure~\ref{fig:cases} through out the evaluation.

\subsection{Evaluation Metrics}
\label{sec:metric}
For the Direction and the Counting categories, for fair evaluation, we adopt both pair-wise accuracy \textit{ACC$_p$} and query-wise accuracy \textit{ACC$_q$} as introduced by \citet{fu2023mme} and \citet{luo2024codis} instead of simple accuracy. For each question, we task models twice with different prompts. \textit{ACC$_p$} only counts when models correctly response to both prompts, while \textit{ACC$_q$} is the averaged score of the two responces. For these different prompts regarding the same question, we vary only the order of images by preserving the clockwise order but changing the beginning view, so that the answer to the question remains unchanged. For example, if original image sequence begins with the northern image, the modified sequence could start from the east, south or west. 

For the Rotation category, some of the instances does not cover the entire space, and the such a modification of input image sequence will inevitable disrupt the comprehension of the continuous space. Similarly, for the Planning category, considering that the original input sequence features the front view as the first image and this manner inherently matches the thinking pattern when models take actions. Therefore, for these two categories, we apply only a single context for each instances during evaluation, and report accuracy, \textit{ACC}, as final scores.

\subsection{Models}

We accessed three proprietary models, including GPT-4o~\citep{gpt-4o}, Gemini-1.5-pro~\citep{reid2024gemini} and Claude-3.7-sonnet~\citep{TheC3}, along with 16 widely used open-source VLMs accepting multiple images simultaneously. Among the open-source models, four have more than 70B parameters, while the others range in size from 2B to 13B. We also enable the evaluation for powerful single-image models, such as LLaVA-1.6~\citep{liu2023llava} and LLaMA-3.2~\citep{dubey2024llama3herdmodels}, for further investigation. For fair comparison, we design a single-image pipeline for some our designed tasks and conduct evaluations on both single-image and multi-image models in Section~\ref{sec:sigimage}.

\subsection{Results}
The overall results on our benchmark is reported in Table~\ref{tab:main_res}. We derive three main findings from this table: 1. \textit{The performance of proprietary models and $>$70B open-source models largely varies.} As expected, most proprietary models and $>$70B open-source models marginally outperform the evaluated open-source models. However, there are significant dependencies among these models, where Claude-3.5-sonnet achieves an average score of 49.37\%, InternVL2\_5-78B achieves 45.50\%, while LLaVA-OneVision-72B only presents an average score of 29.17\%, even lower than more than half of the $<$13B open-source models. 2. \textit{Proprietary models do not equally perform the best across all the tasks.} Proprietary models fail to beat open-source models on tasks including DIR-Obj and CNT, where $>$70B open-source models exhibit stronger performance. 3. \textit{The ROT-Ang task is quite challanging for open-source models}. It is noticeable that for the ROT-Ang task, all open-source models achieve similar results near 50\%, which are almost equal to random selection. Such result is caused by the fixed predictions over all different context and images input. We will elaborately investigate such phenomena in Section~\ref{sec:analysis-ang}. 

We conduct manual evaluation on all instances to assess the reasonability and difficulty of CoSpace and results are provided in Appendix~\ref{sec:humaneval}. Results show that human performance excels the best models across all tasks. Human annotators achieved an average \textit{ACC$_q$} of 86.25\%, which is significantly higher than that of Claude-3.7-sonnet (59.56\%).

\section{Analysis}
\label{sec:analysis}

\begin{table}[t]
    \centering
    \scriptsize
    \begin{tabular}{@{\hspace{0.05cm}}l@{\hspace{0.05cm}}|@{\hspace{0.05cm}}c@{\hspace{0.05cm}}|@{\hspace{0.05cm}}c@{\hspace{0.05cm}}|@{\hspace{0.05cm}}c@{\hspace{0.05cm}}|c|c}
        \toprule
        Models & \multicolumn{1}{c|@{\hspace{0.05cm}}}{DIR-Rec} & \multicolumn{1}{c|@{\hspace{0.05cm}}}{DIR-Obj} & \multicolumn{1}{c|@{\hspace{0.05cm}}}{CNT} & \multicolumn{1}{@{\hspace{0.05cm}}c@{\hspace{0.05cm}}|@{\hspace{0.05cm}}}{PLA-QA} & \multicolumn{1}{@{\hspace{0.05cm}}c@{\hspace{0.05cm}}}{PLA-Dec} \\
        \midrule\noalign{\vskip -3pt}
        \multicolumn{6}{c}{\cellcolor[HTML]{EFEFEF} \scriptsize\textit{Proprietary Models}} \\
        \noalign{\vskip 1pt}
        Claude-3.7-sonnet & 36.36\hphantom{$^*$} & 31.17\hphantom{$^*$} & 37.20\hphantom{$^*$} & 44.40\hphantom{$^*$} & 44.50\hphantom{$^*$} \\
        Gemini-1.5-pro & 37.20\hphantom{$^*$} & 37.20\hphantom{$^*$} & 27.50\hphantom{$^*$} & 38.03\hphantom{$^*$} & 36.15\hphantom{$^*$} \\
        GPT-4o & 37.20\hphantom{$^*$} & 41.20\hphantom{$^*$} & 39.00\hphantom{$^*$} & 45.07\hphantom{$^*$} & 39.91\hphantom{$^*$} \\
        \midrule\noalign{\vskip -3pt}
         \multicolumn{6}{c}{\cellcolor[HTML]{EFEFEF}
         \scriptsize\textit{$>$70B Open-source Models}} \\
         \noalign{\vskip 1pt}
         InternVL2\_5-78B & 29.60\hphantom{$^*$} & 52.80\hphantom{$^*$} & 44.00\hphantom{$^*$} & 34.27$^*$ & 36.15\hphantom{$^*$} \\
         Qwen2-VL-72B & 30.40\hphantom{$^*$} & 47.60\hphantom{$^*$} & 46.00$^*$ & 37.09\hphantom{$^*$} & 33.33\hphantom{$^*$} \\
         InternVL2-76B & 25.20\hphantom{$^*$} & 38.80\hphantom{$^*$} & 52.00$^*$ & 25.35\hphantom{$^*$} & 25.35\hphantom{$^*$} \\
         LLaVA-OneVision-72B & 20.40$^*$ & 43.60\hphantom{$^*$} & 22.50\hphantom{$^*$} & 36.62$^*$ & 32.39$^*$ \\
        \midrule\noalign{\vskip -3pt}
        \multicolumn{6}{c}{\cellcolor[HTML]{EFEFEF} \scriptsize\textit{$<$13B Open-source Models}} \\
        \noalign{\vskip 1pt}
        MiniCPM-V 2.6 & 22.80\hphantom{$^*$} & 40.00\hphantom{$^*$} & 37.00\hphantom{$^*$} & 39.44\hphantom{$^*$} & 26.76\hphantom{$^*$} \\
        Qwen2-VL-7B & 19.60\hphantom{$^*$} & 35.20\hphantom{$^*$} & 40.50\hphantom{$^*$} & 34.27\hphantom{$^*$} & 23.00\hphantom{$^*$} \\
        Mantis-8B & 34.80$^*$ & 30.80\hphantom{$^*$} & 42.50$^*$ & 31.92\hphantom{$^*$} & 29.11$^*$ \\
        InternVL2-8B & 31.20$^*$ & 36.40\hphantom{$^*$} & 42.50$^*$ & 22.54\hphantom{$^*$} & 26.76\hphantom{$^*$} \\
        VILA1.5-8B & 33.60\hphantom{$^*$} & 29.20$^*$ & 42.00\hphantom{$^*$} & 19.72\hphantom{$^*$} & 39.91\hphantom{$^*$} \\
        Idefics3-8B & 37.60$^*$ & 37.20\hphantom{$^*$} & 32.50\hphantom{$^*$} & 23.94\hphantom{$^*$} & 24.88\hphantom{$^*$} \\
        LLaVA-OneVision-7B & 17.20\hphantom{$^*$} & 38.00$^*$ & 35.00\hphantom{$^*$} & 30.05$^*$ & 27.70$^*$ \\
        Phi-3.5-vision & 26.00$^*$ & 34.40$^*$ & 36.50\hphantom{$^*$} & 31.46\hphantom{$^*$} & 24.41\hphantom{$^*$} \\
        Brote-IM-XXL & 34.40$^*$ & 29.60\hphantom{$^*$} & 29.50\hphantom{$^*$} & 29.58$^*$ & 17.84\hphantom{$^*$} \\
        LongVA-7B & 16.00\hphantom{$^*$} & 31.60\hphantom{$^*$} & 28.50$^*$ & 17.84\hphantom{$^*$} & 28.64\hphantom{$^*$} \\
        Mono-InternVL-2B & 18.80\hphantom{$^*$} & 33.60$^*$ & 15.50$^*$ & 25.35\hphantom{$^*$} & 22.07\hphantom{$^*$} \\
        mPLUG-Owl3-7B & 26.40$^*$ & 29.20\hphantom{$^*$} & 29.50$^*$ & 29.11$^*$ & 32.39$^*$ \\
        \bottomrule
    \end{tabular}
    \vspace{-1em}
    \caption{Results for disrupted image orders. \textit{ACC$_q$} is reported for Direction and Counting tasks and \textit{ACC} is reported for the Planning task. $^*$ indicates the performance is higher than that in Table 2.}
    \label{tab:res_hard}
    \vspace{-1.5em}
\end{table}

\subsection{Disrupted Image Order}

Considering the spatial relationship between input images of our CoSpace, the order of input images can make significant difference on the understanding of the space, especially when the order being thoroughly disrupted. Based on this speculation, we randomly shuffled the input sequence, and provided the directional orientation only of the first image for the model, with directions of the rest images to be inferred by the model itself. We conducted evaluation under such setting on three tasks: Direction, Counting and Planning, and the results are illustrated in Table~\ref{tab:res_hard}.

Surprisingly, the performance of $<$13B models under disrupted image order was not always lower than the basic setting. In fact, half of the $<$13B open-source models achieved a higher score in the DIR-Rec task under disrupted setting. Additionally, for other tasks, an average of three to five open-source models performed slightly better with disrupted image order. Nevertheless, similar phenomenon was not observed for proprietary models and most of the $>$70B open-source models, except LLaVA-OneVision-72B. Also, the evaluated proprietary models similarly underperformed across all tasks under disrupted setting, which aligns more closely with our expectations. For $>$70B open-source models, three of four outperformed on only one task against basic setting, with LLaVA-OneVision-72B producing a completely opposite result. Considering the such phenomenon, we conclude that the results for $<$13B open-source models are partly due to the instability of predictions. Moreover, we acknowledge that the impact of disrupting the image order may not be as significant for models as it is for human, exhibiting the discrepancy between models and human when perceiving continuous visual space via multiple images. 

\subsection{Performance Consistency}

\begin{figure}[]
    \centering
    \hspace*{-0.3cm}
    \includegraphics[scale=.29]{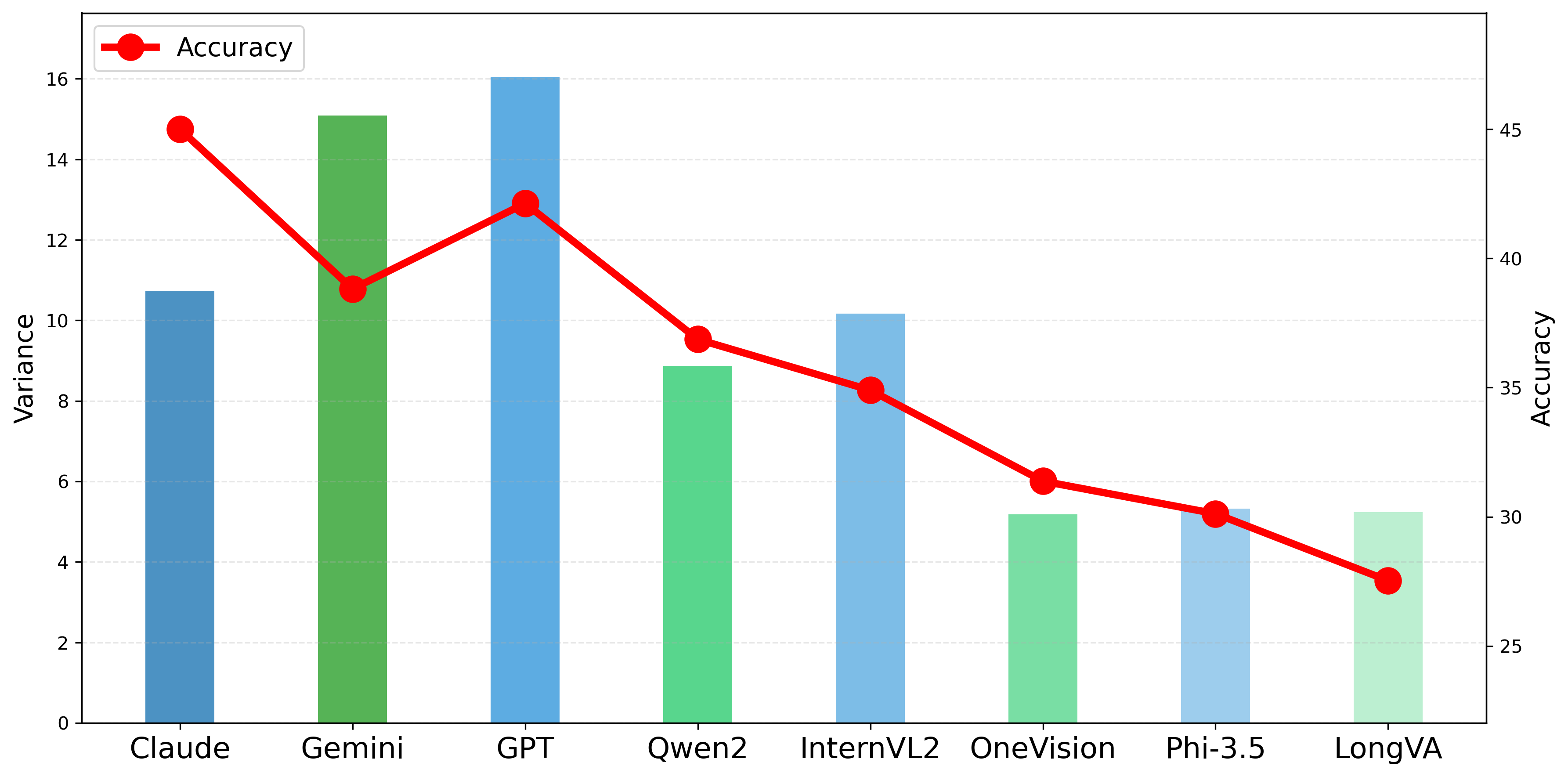}
    \vspace{-2.0em}
    \caption{Variance in \textit{ACC$_q$} of the average on the Direction and Counting categories for eight selected models. Model abbreviations are used for simplicity. Accuracy in the figure represents the average of \textit{ACC$_q$} over two categories.}
    \vspace{-0.5em}
    \label{fig:variance}
\end{figure}

We mentioned in Section~\ref{sec:metric} that for the Direction and Counting categories, both \textit{ACC$_q$} and \textit{ACC$_p$} are adopted as evaluation metrics. We present the gap between \textit{ACC$_q$} and \textit{ACC$_p$}, i.e. variance in accuracy, in Figure ~\ref{fig:variance}. A smaller variance indicates that a model responds more consistently to input changes, reflecting better performance consistency.
As shown, all models exhibit relatively large variance, with proprietary models (Claude-3.7-sonnet, Gemini-1.5-pro, GPT-4o) displaying even higher variances than open-source models, suggesting room for improvement. Such result highlights the importance of introducing \textit{ACC\(_p\)}.

\begin{table}[t]
    \centering
    \scriptsize
    \begin{tabular}{@{\hspace{0.05cm}}l@{\hspace{0.05cm}}|@{\hspace{0.05cm}}c@{\hspace{0.05cm}}c@{\hspace{0.05cm}}c@{\hspace{0.05cm}}|@{\hspace{0.05cm}}c@{\hspace{0.05cm}}c@{\hspace{0.05cm}}c@{\hspace{0.05cm}}c@{\hspace{0.05cm}}c@{\hspace{0.05cm}}}
        \toprule
        Img Order & Claude & Gemini & GPT & Qwen2 & InternVL2 & OneVision & Phi-3.5 & LongVA\\\midrule
        Sequential & 79.58 & 73.64 & 69.91 & 34.25 & 34.78 & 33.33 & 21.62 & 36.59 \\
        Disrupted & 60.71 & 56.79 & 56.25 & 24.66 & 39.58 & 35.94 & 26.87 & 31.58 \\
        \bottomrule
    \end{tabular}
    \vspace{-1em}
    \caption{Results of conditional accuracy on the Planning category for eight selected models. Model names are abbreviated. Open-source models all refer to $<$13B ones.}
    \label{tab:res_rate}
    \vspace{-1.4em}
\end{table}

To further analyze the performance consistency of models, we introduce the metric conditional accuracy in the Planning category. Conditional accuracy measures how often models successfully solve the PLA-Dec question given that they answer the correspondent PLA-QA question correctly, as in PLA-QA task and PLA-Dec task, two respective questions regarding the same set of images typically feature the same object or direction as target. Results in Tabel~\ref{tab:res_rate} reveal that open-source models exhibit a greater gap in conditional accuracy against proprietary models than on the Direction and Counting categories. While proprietary models make more consistent decisions based on correct detection, the gap remains non-negligible. 

Above all, we argue that maintaining performance consistency is rather important for the further improvement of models, as it is closely related to robustness and rather essential for more complicated tasks.

\begin{table}[t]
    \centering
    \scriptsize
    \begin{tabular}{@{\hspace{0.05cm}}l@{\hspace{0.05cm}}|@{\hspace{0.05cm}}c@{\hspace{0.05cm}}|@{\hspace{0.05cm}}c@{\hspace{0.05cm}}|@{\hspace{0.05cm}}c@{\hspace{0.05cm}}|@{\hspace{0.05cm}}c@{\hspace{0.05cm}}|@{\hspace{0.05cm}}c@{\hspace{0.05cm}}|@{\hspace{0.05cm}}c@{\hspace{0.05cm}}}
        \toprule
        Models & \multicolumn{1}{@{\hspace{0.05cm}}c@{\hspace{0.05cm}}|@{\hspace{0.05cm}}}{DIR-Rec} & \multicolumn{1}{@{\hspace{0.05cm}}c@{\hspace{0.05cm}}|@{\hspace{0.05cm}}}{DIR-Obj} & \multicolumn{1}{c|@{\hspace{0.05cm}}}{CNT} & \multicolumn{1}{@{\hspace{0.05cm}}c@{\hspace{0.05cm}}|@{\hspace{0.05cm}}}{PLA-QA} & \multicolumn{1}{@{\hspace{0.05cm}}c@{\hspace{0.05cm}}|@{\hspace{0.05cm}}}{PLA-Dec} & AVG  \\
        \midrule\noalign{\vskip -3pt}
        \multicolumn{7}{c}{\cellcolor[HTML]{EFEFEF} \scriptsize\textit{Models with only one image in the input}} \\
        \noalign{\vskip 1pt}
        LLaVA-v1.6 & 27.60\hphantom{$^*$} & 32.00\hphantom{$^*$} & 21.00\hphantom{$^*$} & 21.60\hphantom{$^*$} & 25.82\hphantom{$^*$} & 25.60\hphantom{$^*$} \\
        LLama-3.2-11B & 29.20\hphantom{$^*$} & 28.80\hphantom{$^*$} & 28.50\hphantom{$^*$} & 17.84\hphantom{$^*$} & 16.90\hphantom{$^*$} & 24.25\hphantom{$^*$} \\
        \midrule\noalign{\vskip -3pt}
        \multicolumn{7}{c}{\cellcolor[HTML]{EFEFEF} \scriptsize\textit{Models with multi-image perception ability}} \\
        \noalign{\vskip 1pt}
        InternVL2-8B & 36.40$^*$ & 34.80\hphantom{$^*$} & 27.00\hphantom{$^*$} & 33.80$^*$ & 31.92$^*$ & 32.78\hphantom{$^*$} \\
        MiniCPM-V 2.6 & 23.20\hphantom{$^*$} & 27.20\hphantom{$^*$} & 26.00\hphantom{$^*$} & 31.92\hphantom{$^*$} & 32.39$^*$ & 28.14\hphantom{$^*$} \\
        Phi-3.5-vision & 18.40\hphantom{$^*$} & 36.00$^*$ & 28.00\hphantom{$^*$} & 31.46\hphantom{$^*$} & 23.00\hphantom{$^*$} & 27.37\hphantom{$^*$} \\
        LLaVA-OneVision & 26.80$^*$ & 32.80\hphantom{$^*$} & 16.00\hphantom{$^*$} & 22.07\hphantom{$^*$} & 24.41\hphantom{$^*$} & 24.42\hphantom{$^*$} \\
        \bottomrule
    \end{tabular}
    \vspace{-1em}
    \caption{Results for single image pipeline. We report \textit{ACC$_q$} for Direction and Counting tasks and \textit{ACC} for Planning tasks. $^*$ indicates the performance is higher than that in Table~\ref{tab:main_res}.}
    \label{tab:res_sig}
    \vspace{-1.5em}
\end{table}

\subsection{Single Image Pipeline}
\label{sec:sigimage}

We presented in Section~\ref{sec:spaceperception} that the understanding of continuous visual space inherently relies on multiple images as input to represent the whole space. Also, recent work~\citep{wang2024actiview} explored to construct a pipeline that can utilize caption and memory to transform the multi-image task into single image format. Similarly, we developed a single image pipeline for the Direction, Counting and Planning categories in our benchmark. For the Rotation category, as the single image input will inevitably compromise the uniform understanding of continuous visual space and the relationship between images can hardly be replaced by mere captions, we consider it not suitable for the single image pipeline. For evaluation, we selected LLaVA-v1.6~\citep{liu2023llava} and Llama-3.2-Vision~\citep{dubey2024llama3herdmodels} as the representatives of single image models. We also selected some multi-image models and conducted the single image evaluation.

Results of the single image pipeline are reported in Table~\ref{tab:res_sig}. Overall, the multi-image models underperformed under this setting compared with the multi-image pipeline. However, in the Direction-Rec and Planning-Dec task, two of four models achieve a higher accuracy, which is far beyond our expectations. We conclude that with the effective summarization of captions, models can still have an understanding of visual clues and sometimes the complete of all images can even add to the performance. Also, three of the four multi-image models outperform the single-image models on the average accuracy, indicating that models with inherent multi-image capabilities boast an overall better understanding of the visual clues and information extracted from the effectively multi-image scene. Based on this observation, the importance of empowering models with multi-image capabilities needs again to be emphasized.

\subsection{Investigation on Angle-related Reasoning}
\label{sec:analysis-ang}

Table~\ref{tab:main_res} shows that all the assessed open-source models achieve near-half \textit{ACC}s on the Rotation-Angle (ROT-Ang) task. 
After checking the responses of different models, we found out that these models tend to produce the same answer across all the queries, e.g. ``\textit{A}'' for all the 300 questions in this task. Therefore, due to the even distribution of options, these models achieve the similar near-half \textit{ACC}.

To discover the reason behind such phenomenon, we asked models to generate the rationale when answering the questions in the ROT-Ang task and observed some of these models can follow the instruction to generate rationales. For example, InternVL2-8B no longer always selected the same option and consequently correctly answered some question that it originally failed. However, we also noticed that it tended to identify the whole surrounding covered by the images, thus perceiving the sum of turning angles as 360 degrees. Therefore, it frequently output 90 degrees as the final answer, even if 90 degrees not included in the options. In contrast, proprietary models like Claude-3.7-sonnet could firstly calculate the sum of covered angles across all images based on the two give options and then decided which answer better matched the scene in given images.

Though all the assessed open-source models fail the Rotation-Angle task, we still need to underline the importance of this task and the related capability to correctly identify the rotation in the space. Such capability lays the foundation for more advanced tasks in practical use and has already been boasted by some of the proprietary models.

\section{Conclusion}
\label{sec:conclusion}

Recognizing the gap between previous work and more advanced spatial understanding, we propose the concept of continuous space perception and present a novel benchmark CoSpace that comprises four categories and seven tasks. CoSpace covers both fundamental spatial comprehending abilities and more practical and commonly-used capabilities to comprehensively assess continuous space perception ability for VLMs. We evaluated three widely used proprietary models and sixteen most advanced open-source models and conducted detailed analysis for noticeable results in the experiments. Results reveal a gap between the assessed open-source and proprietary models, especially for performance consistency and in rotation-based perception.

\section*{Acknowledgments}

This work is supported by the National Key R\&D Program of China (2022ZD0160502), the National Natural Science Foundation of China (No. 62276152), and funding from Wuxi Research Institute of Applied Technologies, Tsinghua University under Grant 20242001120. We acknowledge Yan Liu, Jidong Chen, Jianyang Liu, Shuangyu Li and Shu Chi for their effort during the process of data annotation and human evaluation.
{
    \small
    \bibliographystyle{ieeenat_fullname}
    \bibliography{main}
}

\clearpage
\maketitlesupplementary

\appendix
\section{Details for Data Collection and Annotation}

\paragraph{Image Collection.} Our benchmark comprises 2,918 images, including 2,302 outdoor images from publicly available APIs and 616 indoor images from the simulator platform. This section reports details for image collection.

To collect required outdoor images, we utilize two different API interfaces from Baidu Map API\footnote{\url{https://lbsyun.baidu.com/}}: the Place API\footnote{\url{https://lbsyun.baidu.com/faq/api?title=webapi/guide/webservice-placeapi}} and the Panorama API\footnote{\url{https://lbsyun.baidu.com/faq/api?title=viewstatic}}. For Place API, we first provide some keywords, including 8 keywords for cities and 12 keywords for places as follows:

\begin{itemize}
    \item \textbf{City:} Beijing, Shanghai, Guangzhou, Shenzhen, Hangzhou, Nanjing, Chengdu, Chongqing.
    \item \textbf{Place:} school, hospital, museum, park, library, mall, cinema, railway station, airport, stadium, supermarket.
\end{itemize}
The Place API is responsible for providing locations, in the form of latitude and longitude, e.g. \textit{\{``lat'': 40.099567, ``lng'': 116.515935\}}, of all the combinations of the given keywords, such as ``schools in Beijing'' and ``parks in Chongqing''. Then, given the latitude and longitude of a location, the the Panorama API returns the image set (views of four orientations, north, east, south, and west) of the corresponding location following our defined continuous visual space format as elaborated.

We utilized Habitat-Sim platform~\citep{habitat19iccv, szot2021habitat, puig2023habitat3} to collect images of indoor scenes from the simulator environment. This platform provides an interface to instruct a virtual robot to explore the surroundings and capture images of its views. Iteratively, we place the robot at random positions and capture views of four directions to the robot, including front, right, back and left, making them to form a continuous visual space as illustrated. 

In total, the APIs and the platform generated 2,883 and 6,000 images respectively. We then manually filtered the images accessed from HM3D dataset and finally preserve 800 images. The filtering requirements are as follows: as a large number of images collected from HM3D are shot in the same scene and contain similar visual information and we require that there should exist notable difference between different sets of images to avoid duplication.

\paragraph{Data Annotation.}
After collecting and filtering the images, we follow a two-phase paradigm for annotation: firstly utilizing GPT-4o to generate questions, and then asking human annotators to provide the groundtruth answers for the generated questions. The prompt templates for question generation using GPT-4o are as following:

\begin{tcolorbox}[breakable, title=Prompt Templates Used for Data Annotation]
    \small
    \textbf{System Prompt for Direction and Counting Category:} \\
    You are a data curation engineer and need to generate some **question**-**answer** pairs according to the requirements and given images. \\
    You will be given an example, which includes a set of images and a perfect **question**-**answer** pair related to the images set. \\ \\
    You should follow the given example and generate **question**-**answer** pairs for another images set. \\
    You can generate one or more **question**-**answer** pairs. These pairs should be practical, accurate according to the images.\\ \\
    Give your output in the following JSON format: \\
     \text{[}
        \hphantom{1111} \\
        \hphantom{1111}\{ \\
            \hphantom{11111111}"question": "some text", \\
            \hphantom{11111111}"answer": "some text", \\
        \hphantom{1111}\}, \slash\slash\hphantom{1}question-answer pair 1 \\
        \hphantom{1111}\{ \\
            \hphantom{11111111}"question": "some text", \\
            \hphantom{11111111}"answer": "some text", \\
        \hphantom{1111}\} \slash\slash\hphantom{1}question-answer pair 2 \\
    \text{]}
    \vspace{0.7em}\hrule\vspace{0.7em}
    \textbf{System Prompt for Planning Category:} \\
    You are a data curation engineer and need to generate some **question**-**answer** pairs according to the requirements and given images. \\
    You will be given an example, which includes a set of images and a perfect **question**-**answer** pair related to the image set. \\ \\
    I want you to follow the example and generate some similar **question**-**answer** pairs for another images set. \\
    You can generate one **question**-**answer** pair or more, as long as you can ensure the quality and correctness of your output. \\ \\
    Give your output in the following JSON format: \\
     \text{[}
        \hphantom{1111} \\
        \hphantom{1111}\{ \\
            \hphantom{11111111}"question-qa": "some text", \\
            \hphantom{11111111}"answer-qa": "some text", \\
            \hphantom{11111111}"question-dec": "some text", \\
            \hphantom{11111111}"answer-dec": "some text", \\
        \hphantom{1111}\}, \slash\slash\hphantom{1}question-answer pair 1 \\
        \hphantom{1111}\{ \\
            \hphantom{11111111}"question-qa": "some text", \\
            \hphantom{11111111}"answer-qa": "some text", \\
            \hphantom{11111111}"question-dec": "some text", \\
            \hphantom{11111111}"answer-dec": "some text", \\
        \hphantom{1111}\} \slash\slash\hphantom{1}question-answer pair 2 \\
    \text{]} \\
    You should notice that there is a **question**-**answer** pair for qa and **question**-**answer** pair for dec in each output item. I want them to focus on the same object. The qa question should be about where a certain object is and the dec question should be about how to fetch that specific thing or something else related to that. \\
    The question for the dec question can only be composed of the following actions: go ahead, turn left, turn right, turn back. So the accepted choices and answer are like "turn right and go ahead" or "go ahead and turn left".
    \vspace{0.7em}\hrule\vspace{0.7em}
    \textbf{User Prompt:} \\
    \#\# Task description \\ \\
    Now you need to generate some space and directions related **question**-**answer** pairs. \\
    You are given eight images. The first four belong to the example image set, while others belong to the test image set.
    The **question**-**answer** given in the example is based on the first four images and the **question**-**answer** pairs you generate should be based on the test images set.
    Also, you need to set this task as the form choice as is shown in the example. \\ \\
    \#\# Image explanation \\ \\
    The images in this task are arranged in the following sequence: \\
    All the images in a set are shot in the same scene, but of four directions. \\
    The **first** image is facing **north**, towards the **front** side. \\
    **Second** facing **east**, towards the **right** side. \\
    **Third** facing **south**, towards the **back** side. \\
    **Fourth** facing **west**, towards the **left** side. \\
    Also, there are some overlap between the adjacent images. \\
    
    \#\# Example task \\ \\
    \slash\slash\hphantom{1}A given question-answer pair as in-context example
\end{tcolorbox}

We included a one-shot in-context example in the prompt to help GPT-4o better understand the requirements for question generation. Meanwhile, GPT-4o is required to generate a corresponding answer to the question. However, during manual review, we observed that although the generated questions were appropriate, GPT-4o often produced incorrect answers that did not align with the given images. To this end, we manually annotated all the answers for the generated questions. For each question-answer pairs, an average of two annotators are involved to ensure reliability. After annotation, 2,302 of 2,883 outdoor images and 616 of 800 images acquired from HM3D dataset are left, finally comprising our CoSpace.

\section{Prompt Templates for Evaluation}

We report our prompt templates used in the experiments for evaluation as follows:

\begin{tcolorbox}[breakable, title=Prompt Templates for Evaluation]
    \small
    \textbf{Direction Category:} \\ \\
    You are provided with four images shot in the same scene towards different direction. These images overlap in a certain manner, and are arranged in the following order: \\ \\
    \textbf{\{order\}} \\ \\
    Carefully analyze these images, and answer the following question from the given options. \\
    Question: \textbf{\{question\}}. Options: \textbf{\{options\}}. \\ \\
    You should generate your answer from `A, B, C or D'. Your answer:
    \vspace{0.7em}\hrule\vspace{0.7em}
    \textbf{Counting Category:} \\ \\
    You are provided with four images shot in the same scene towards different direction. These images overlap in a certain manner, and are arranged in the following order: \\ \\
    \textbf{\{order\}} \\ \\
    Carefully analyze these images, and answer the following question. \\
    Question: \textbf{\{question\}}. \\ \\
    You should generate a single number as your answer. Your answer:
    \vspace{0.7em}\hrule\vspace{0.7em}
    \textbf{Rotation-Angle Task:} \\ \\
    You are provided with four images shot in the same scene. They are taken from the same position but towards different directions. For example, after taking the first image, the photographer turns a certain degree clockwise. We denote this degree as the turning degree between two adjacent images. These images are arranged in the following order: they are arranged clockwise and the turning angle between adjacent images are the same. \\ \\
    Also note that the image sequence does not always cover a full 360-degree scene. The covered degree of the image sequence can range from 90 to 360. \\ \\
    Carefully analyze these images, and answer the following question from the given options. \\
    Question: \textbf{\{question\}}. Options: \textbf{\{options\}}. \\ \\
    You should generate your answer from `A or B'. Your answer:
    \vspace{0.7em}\hrule\vspace{0.7em}
    \textbf{Rotation-Difference Task:} \\ \\
    You are provided with five images shot in the same scene at the same position towards different directions. In these five images, four are taken in the following way: after taking the first image, the photographer turns a certain degree clockwise to take the next one and the degree always remains the same. \\ \\
    These images are also arranged as the sequence they are taken. However, the rest one image is shot towards totally different direction and is randomly inserted into the image sequence. \\ \\
    Carefully analyze these images, and answer the following question. \\
    Question: \textbf{\{question\}} \\ \\
    You should generate a single number as your answer, where 1 represents the first image and 5 represents the last image. Your answer:
    \vspace{0.7em}\hrule\vspace{0.7em}
    \textbf{Planning Category:} \\ \\
    You are a human like robot. You can only go straight ahead If you want to walk in the other direction, you need to first turn to the target direction and then move forward. The images are arranged in the following order: \\ \\
    \textbf{\{order\}} \\ \\
    There are two questions for you to answer at the same time. Please carefully analysis your surroundings and answer the following questions: \\
    Question 1: \textbf{\{question\_qa\}} Options: \textbf{\{options\_qa\}}. \\
    Question 2: \textbf{\{question\_dec\}} Options: \textbf{\{options\_dec\}}. \\ \\
    You should generate your answer in a JSON dict containing 2 fields: \\
    \{ \\
    \hphantom{1111}`Answer1': type str, answer to question 1, in the form of `A', `B', `C' or `D', \\
    \hphantom{1111}`Answer2': type str, answer to question 2, in the form of `A', `B', `C' or `D'. \\
    \} \\ \\
    Your response:
\end{tcolorbox}

For each single query, \textbf{\{question\}}, \textbf{\{options\}}, \textbf{\{question\_qa\}}, \textbf{\{question\_qa\}}, \textbf{\{question\_dec\}} and \textbf{\{options\_dec\}} are replaced by query-specific question and options. Also \textbf{\{order\}} are replaced with detailed explanation of the input image order, tailored to different tasks and settings.

However, some of the assessed models do not follow instructions properly, and slightly adapt the templates for them to obtain valid responses. For instance, Mantis-8B and VILA1.5-8B output the single ``A'' as the answer for all queries regardless of the question and options. Moreover, models like Mono-InternVL-2B and Brote-IM-XXL can not follow the instruction to output the required JSON dict in the Planning category. For the evaluation of models that output the same answer for all questions, we use different prompts. For instance, we only maintain question, options and the explanation of order for the evaluation of Mantis-8B, which greatly simplifies the prompt template and leads to the better response. For the models that cannot follow instructions for the Planning category, we adopt the strategy of asking models to response to the PLA-QA task and PLA-Dec task separately.

\section{Discussion on Open-ended Evaluation}
As mentioned, we adopted the form of multiple choices for all tasks except for the ROT-Dif task. Our benchmark typically features tasks in real-world scenarios. In practical applications, models are often required to handle fully open-ended questions without being constrained to a set of predefined choices. However, we chose not to evaluate these tasks in an open-ended setting for the following concerns:

\begin{itemize}
    \item For DIR-Rec and PLA-QA tasks, the answers universally contain fixed directions (e.g. ``south'') or trajectory directions (e.g. ``east to west''). These answers fall in a certain range, meaning that in an open-ended evaluation, models are actually choosing from a fixed and implicit set of options, with the number of options being more than four. Therefore, we conclude these tasks as semi-open-ended, reducing the necessity of conducting fully open-ended evaluations.
    \item Similarly, the PLA-Dec task also features semi-open-ended answers, because the action space for this task is limited to ``go ahead'', ``turn right'', ``turn left'' and ``turn back''. Any final decision needs to be composed of these atomic actions. Consequently, open-ended evaluation can be replaced by providing multiple options and we argue that providing four options is sufficient for the evaluation of the continuous space perception ability.
    \item Open-ended evaluation is not suitable for the DIR-Obj task, which requires models to identify existing objects regarding the given direction. There usually exist more than one objects in a given direction and all these objects should be noted as potential answers, significantly increasing the difficulty and ambiguity of evaluation. Thus, open-ended evaluation is not employed for this task.
\end{itemize}

To summarize, the current evaluation setting can provide us with a comprehensive understanding of the assessed continuous space perception ability and we chose not to implemented open-ended setting for further evaluation.

\section{Human Evaluation}
\label{sec:humaneval}

To assess the difficulty and reasonability of our benchmark, we conducted an extensive human evaluation with each sample tested two times. We provide results for human evaluation in Table~\ref{tab:human}. Concluded from this table, humans achieve significantly higher accuracy compared to the best scores from models for all the tasks except for ROT-Dif, where the best model performance only lags behind by 2.09\%. The superiority of Claude-3.7-sonnet in the ROT-Dif task (93.50\%) lies in the sensitivity inconsistencies within a series of continuous images. In this task, human might overlook subtle inconsistencies, especially when the differences are as small as for ROT-Dif task.

\begin{table}[h]
    \centering
    \resizebox{0.48\textwidth}{!}{
    \begin{tabular}{@{\hspace{0cm}}l@{\hspace{0.1cm}}|@{\hspace{0.1cm}}c@{\hspace{0.1cm}}|@{\hspace{0.1cm}}c@{\hspace{0.1cm}}|@{\hspace{0.1cm}}c@{\hspace{0.1cm}}|@{\hspace{0.1cm}}c@{\hspace{0.1cm}}|@{\hspace{0.1cm}}c@{\hspace{0.1cm}}|@{\hspace{0.1cm}}c@{\hspace{0.1cm}}|@{\hspace{0.1cm}}c@{\hspace{0cm}}}
        \toprule
         & DIR-Rec & DIR-Obj & CNT & ROT-Ang & ROT-Dif & PLA-QA & PLA-Dec \\
        \midrule
        Random & 24.82 & 24.91 & 10.01 & 49.37 & 19.88 & 25.11 & 24.30 \\
        Models & 44.40 & 54.40 & 51.25 & 64.33 & 93.50 & 54.73 & 69.34 \\
        Human & \textbf{82.40} & \textbf{80.20} & \textbf{78.25} & \textbf{96.17} & \textbf{95.59} & \textbf{88.26} & \textbf{82.87} \\
        \bottomrule
    \end{tabular}
    }
    \caption{Results for human evaluation. We report the average accuracy of human annotators, and take the highest accuracy among all assessed models for each task as comparison.}
    \label{tab:human}
\end{table}

\section{Details for Single Image Pipeline}

For single-image models, we convert images into captions to enable the evaluation. In this section we provide the used prompt template and cases for the single image pipeline.

\begin{figure}
    \centering
    \includegraphics[scale=0.67]{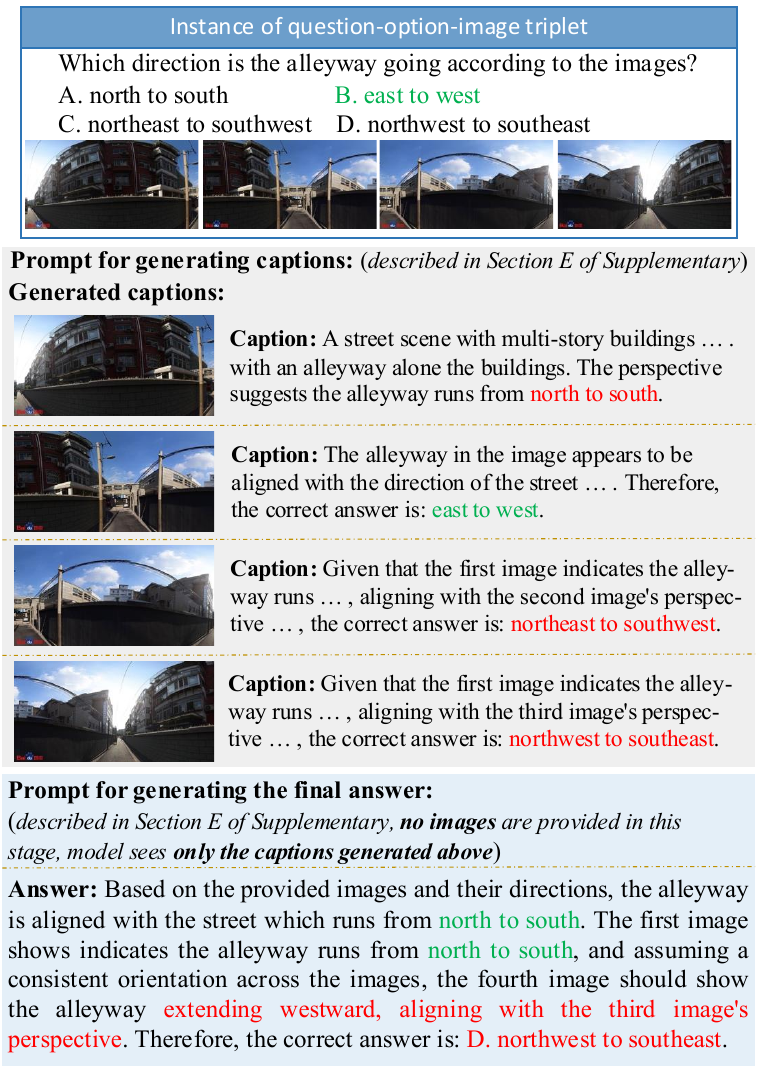}
    \vspace{-2em}
    \caption{Case for single image pipeline. For illustration, we showcase all the images in the figure, \underline{but models can only see one image at the same time}. The responses in this case are all generated by MiniCPM-V 2.6.}
    \label{fig:single}
\end{figure}

\begin{tcolorbox}[breakable, title=Prompt Template Used for Single Image Pipeline]
    \small
    \textbf{Prompts for Generating Captions:} \\ \\
    You are a helpful assistant and is now faced with a task. There are a series of images and several questions related to it. However, you can access to only one image at the same time. Therefore, you need to write down some captions about the image when you view it to help you answer the question. Finally, you will be provided with only the captions you write down when tasked with answering the question. \\ \\
    \textbf{\{query\}} \textcolor{gray}{\slash\slash \hphantom{1}Same as the query for regular evaluation, containing task descriptions, explanations of input images and the question.} \\ \\
    This is the \{first, second, third, fourth\} image. Now you can write down the caption of this image to help you finally answer the question. You should notice that you will not be provided with the images when generating final answer, so the caption should be as detailed as possible. The captions of the past images are listed below: \\ \\
    \textbf{\{captions\}} \textcolor{gray}{\slash\slash \hphantom{1}The captions for the past images. Captions are generated as the order of input images.}
    \vspace{0.7em}\hrule\vspace{0.7em}
    \textbf{Prompts for Generating the Final Answer:} \\ \\
    You are a helpful assistant and is now faced with a task. There are a series of images and several questions related to it. However, you can access to only one image at the same time. Therefore, you need to write down some captions about the image when you view it to help you answer the question. Finally, you will be provided with only the captions you write down when tasked with answering the question. \\ \\
    Your captions of all the images are listed below: \\ \\
    \textbf{\{captions\}} \textcolor{gray}{\slash\slash \hphantom{1}Generated captions for all images.} \\ \\
    Now answer the given question and you should output in the required format. \\ \\
    \textbf{\{query\}} \textcolor{gray}{\slash\slash \hphantom{1}Query containing the task description, explanation for the input images and the question, same as the prompt of the regular setting.}
\end{tcolorbox}

As shown in Figure~\ref{fig:single}, MiniCPM-V 2.6 captured the existence of alleyway in the first image which is facing towards north but mistakenly captioned that the alleyway run from north to south. Actually, the alleyway is running parallel in the first image, standing for the direction of east to west. Though the model correctly identify the direction through the second image, when generating the captions for other two images, it was mistaken by the caption of the first image and finally generate the wrong answer.

\section{Impact of Rationales on the ROT-Ang Task}

\begin{figure}
    \centering
    \includegraphics[scale=0.72]{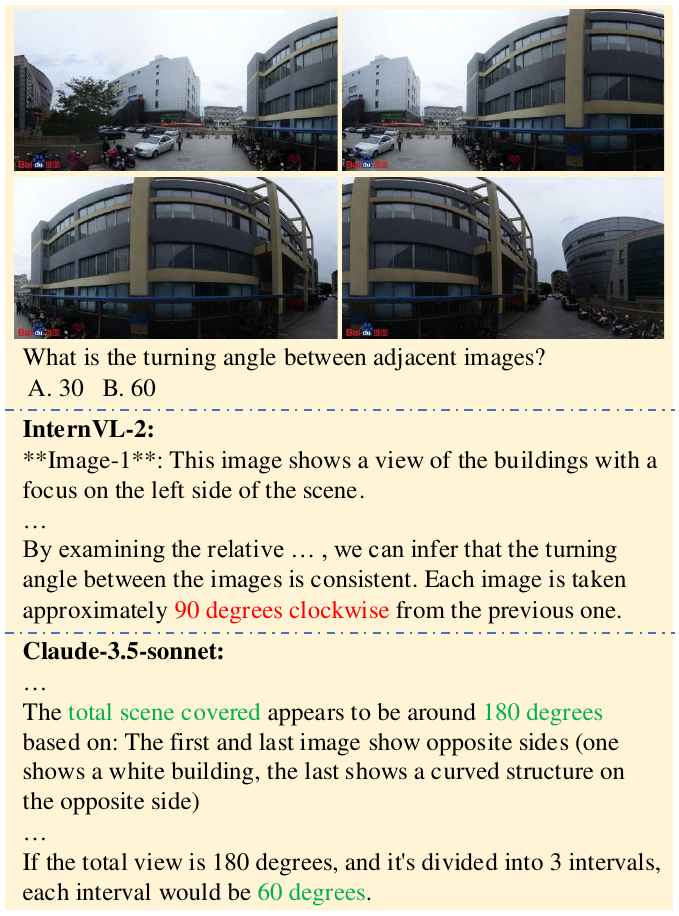}
    \caption{Cases of generated rationales in the Rotation-Angle task.}
    \label{fig:angle}
\end{figure}

We notice that most of the evaluated models fail to response properly for the Rotation-Angle (ROT-Ang) task. To further investigate into this phenomenon, we provide two examples of the rationales respectively generated by InternVL-2 and Claude-3.5-sonnet in the ROT-Ang task. As shown in Figure~\ref{fig:angle}, InternVL-2 outputs brief and generic captions for each image, and mistakenly perceives the turning angle between images as 90 degrees, even if 90 is not included in the options. In contrast, Claude-3.5-sonnet correctly identify the total scene coverage as 180 degrees and successfully recognizes the three equal intervals, which helps it accurately derive the answer of 60 degrees.

\section{Case Study}

\begin{figure*}
    \centering
    \captionsetup{type=figure}
    \includegraphics[width=1.0\textwidth]{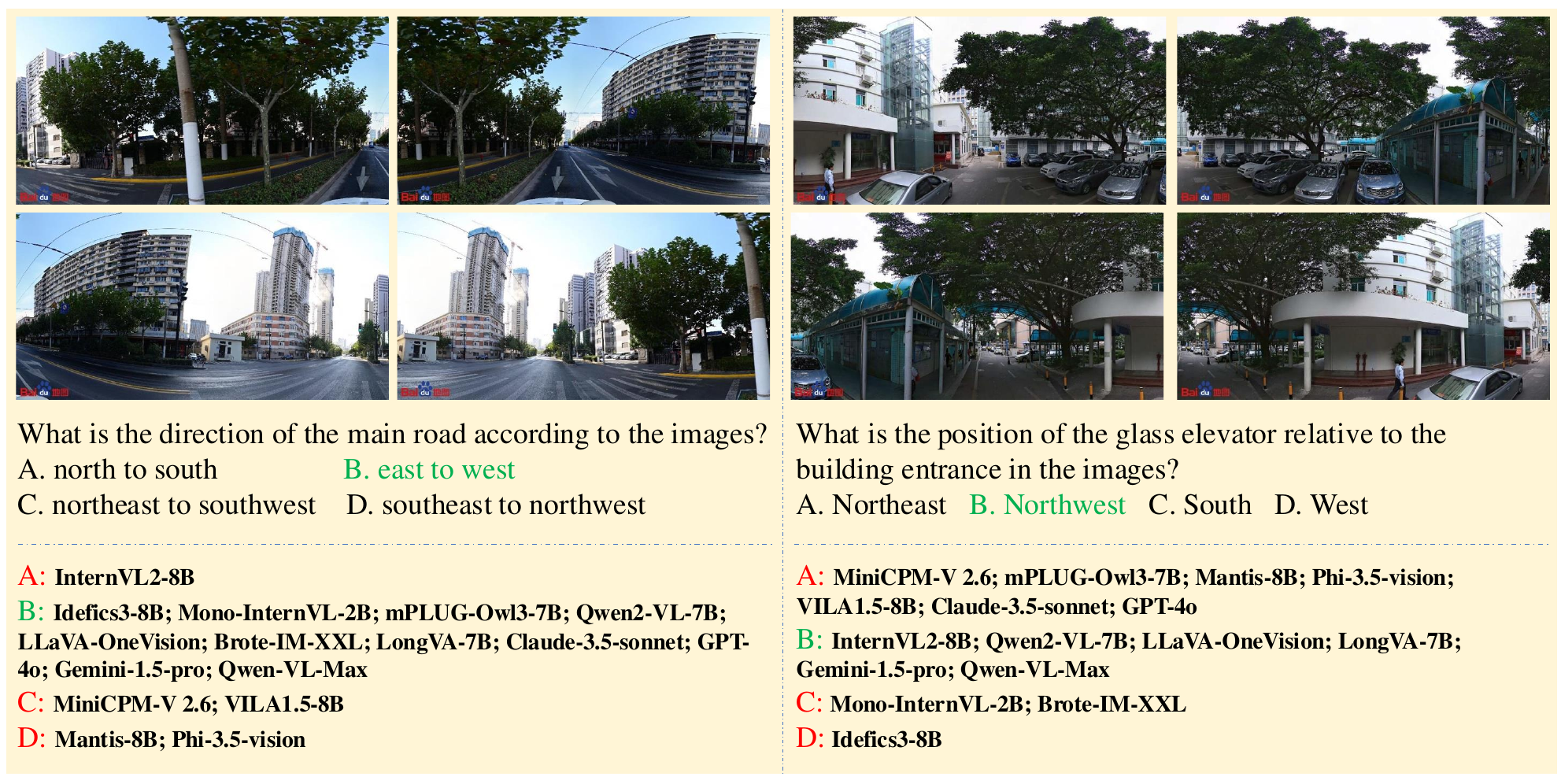}
  \caption{Cases for Direction Recognition task. Images are arranged as the following order: the first image is facing towards north, second facing east, third facing south and fourth facing west.}
  \label{fig:case1}
\end{figure*}

\begin{figure*}
    \centering
    \captionsetup{type=figure}
    \includegraphics[width=1.0\textwidth]{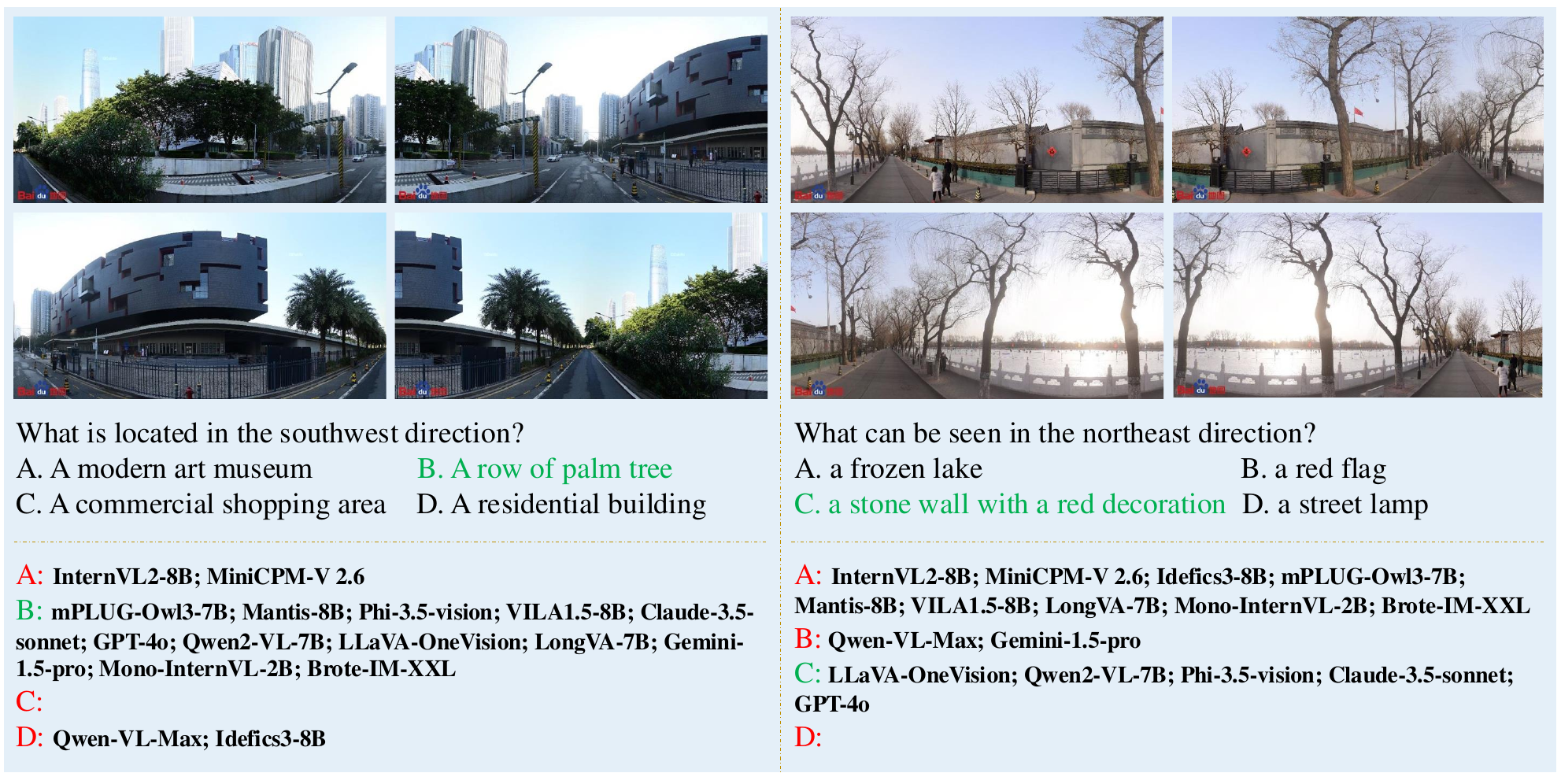}
  \caption{Cases for Direction Object Perception task. Images are arranged as the following order: the first image is facing towards north, second facing east, third facing south and fourth facing west.}
  \label{fig:case2}
\end{figure*}

\begin{figure*}
    \centering
    \captionsetup{type=figure}
    \includegraphics[width=1.0\textwidth]{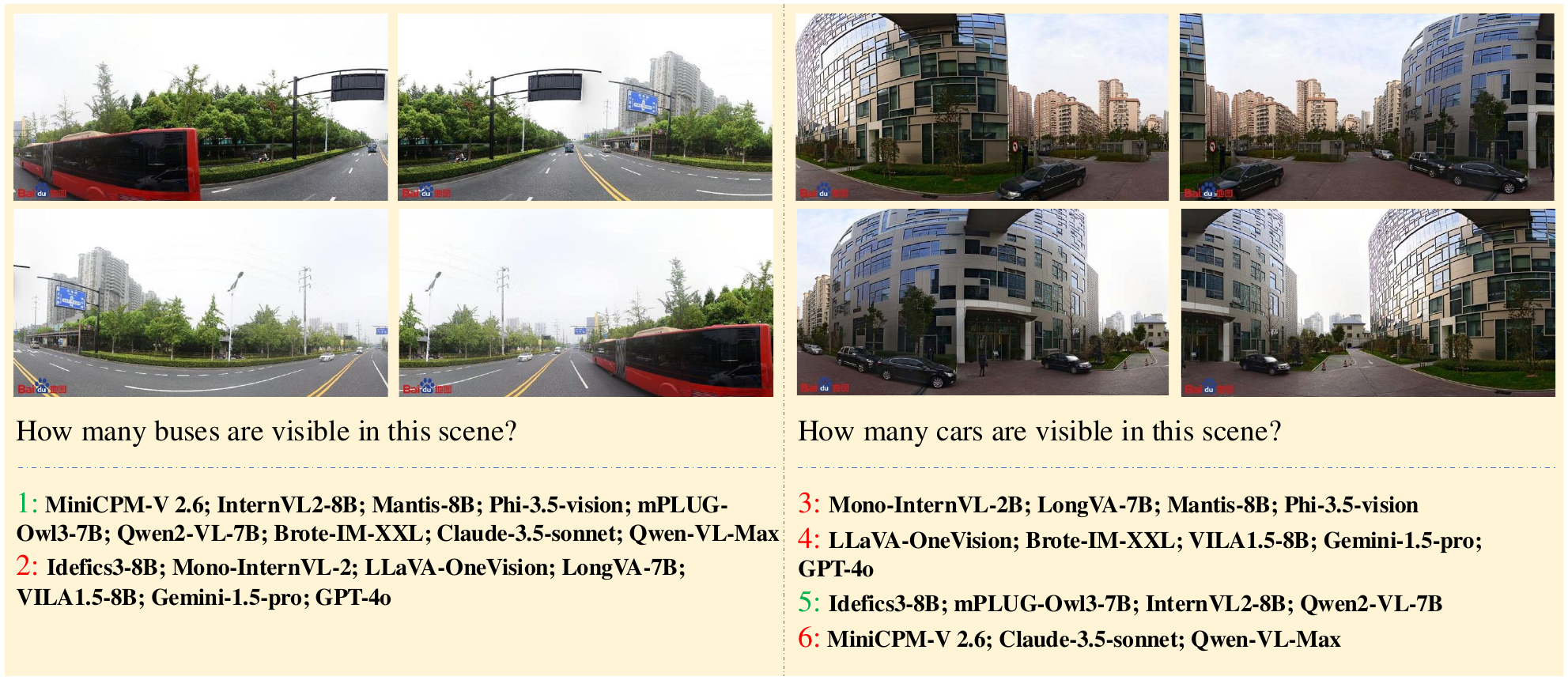}
  \caption{Cases for Counting task. Images are arranged as the following order: the first image is facing towards north, second facing east, third facing south and fourth facing west.}
  \label{fig:case3}
\end{figure*}

\begin{figure*}
    \centering
    \captionsetup{type=figure}
    \includegraphics[width=1.0\textwidth]{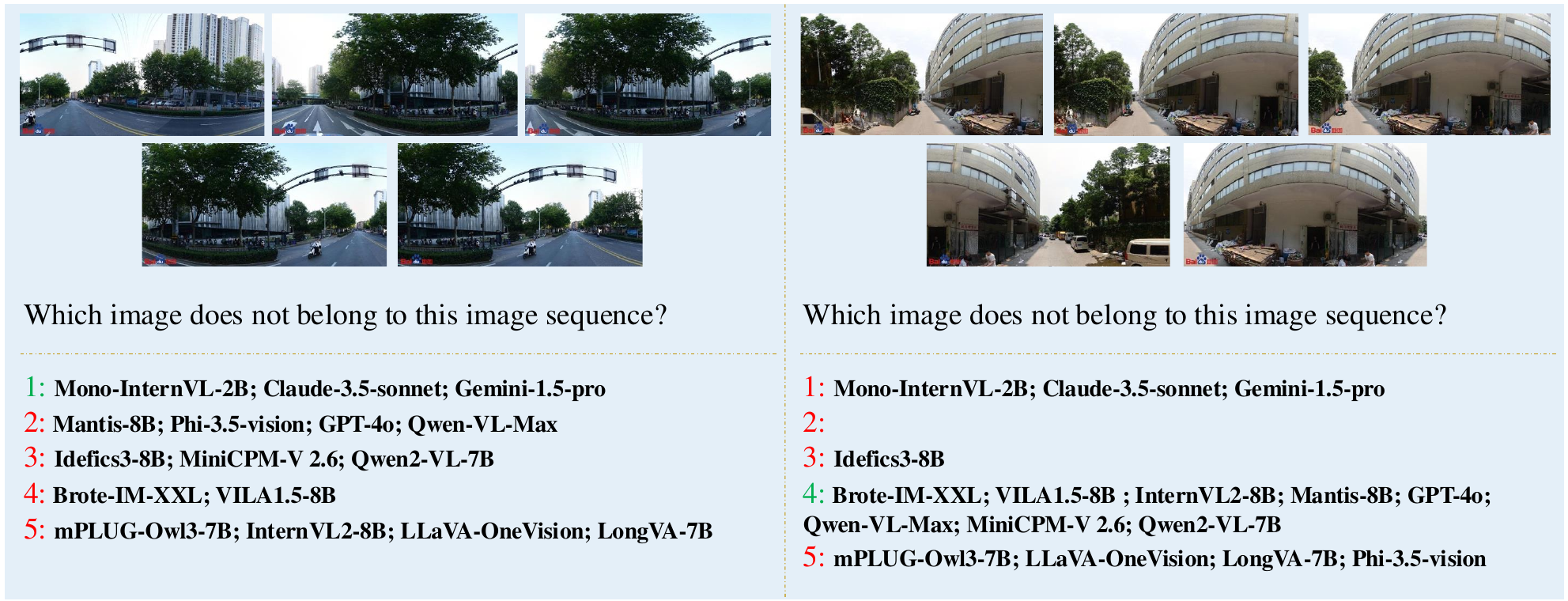}
  \caption{Cases for Direction Rotation Difference task.}
  \label{fig:case4}
\end{figure*}

\begin{figure*}
    \centering
    \captionsetup{type=figure}
    \includegraphics[width=1.0\textwidth]{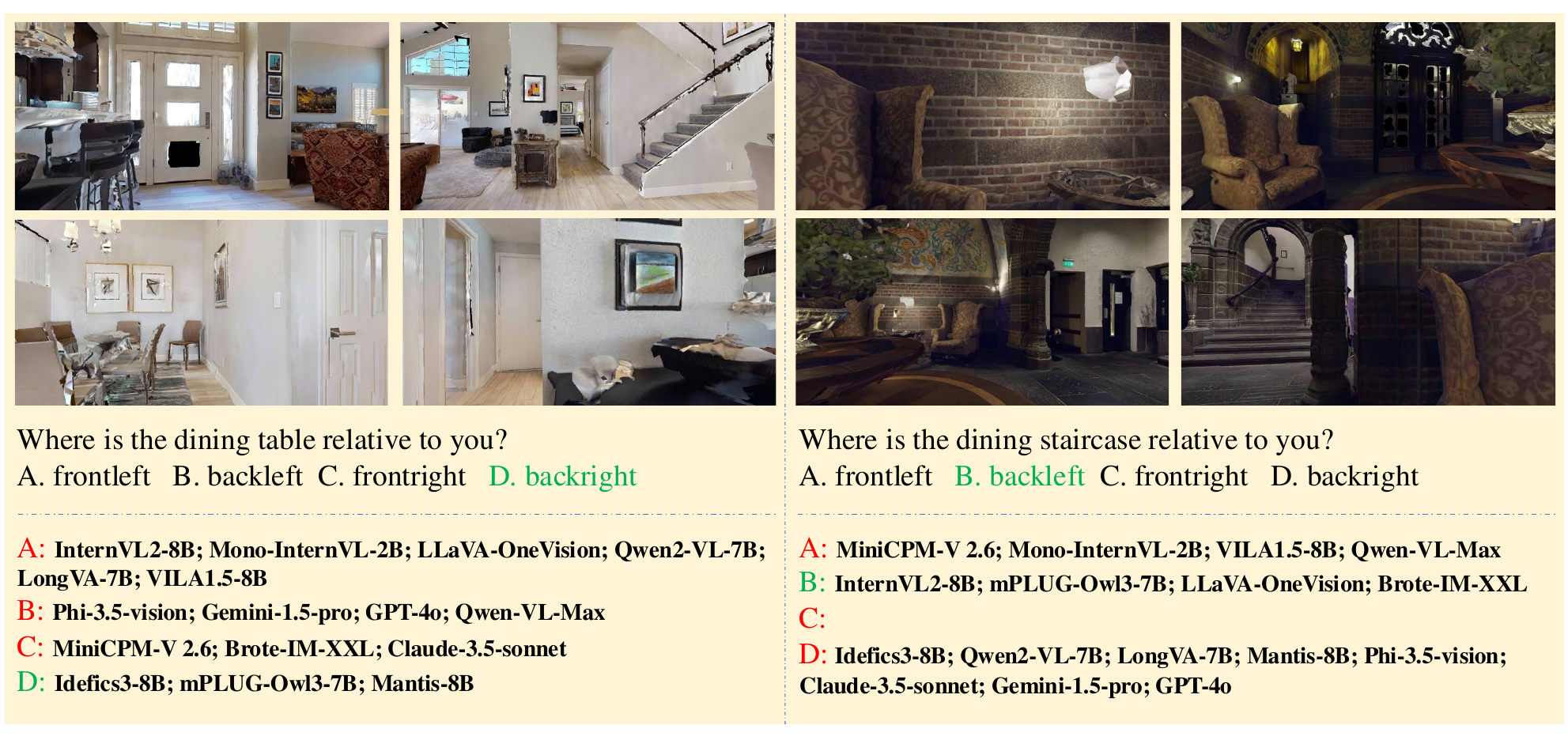}
  \caption{Cases for Planning Question Answering task. Images are arranged as the following order: the first image is facing towards front, second facing right, third facing back and fourth facing left.}
  \label{fig:case5}
\end{figure*}

\begin{figure*}
    \centering
    \captionsetup{type=figure}
    \includegraphics[width=1.0\textwidth]{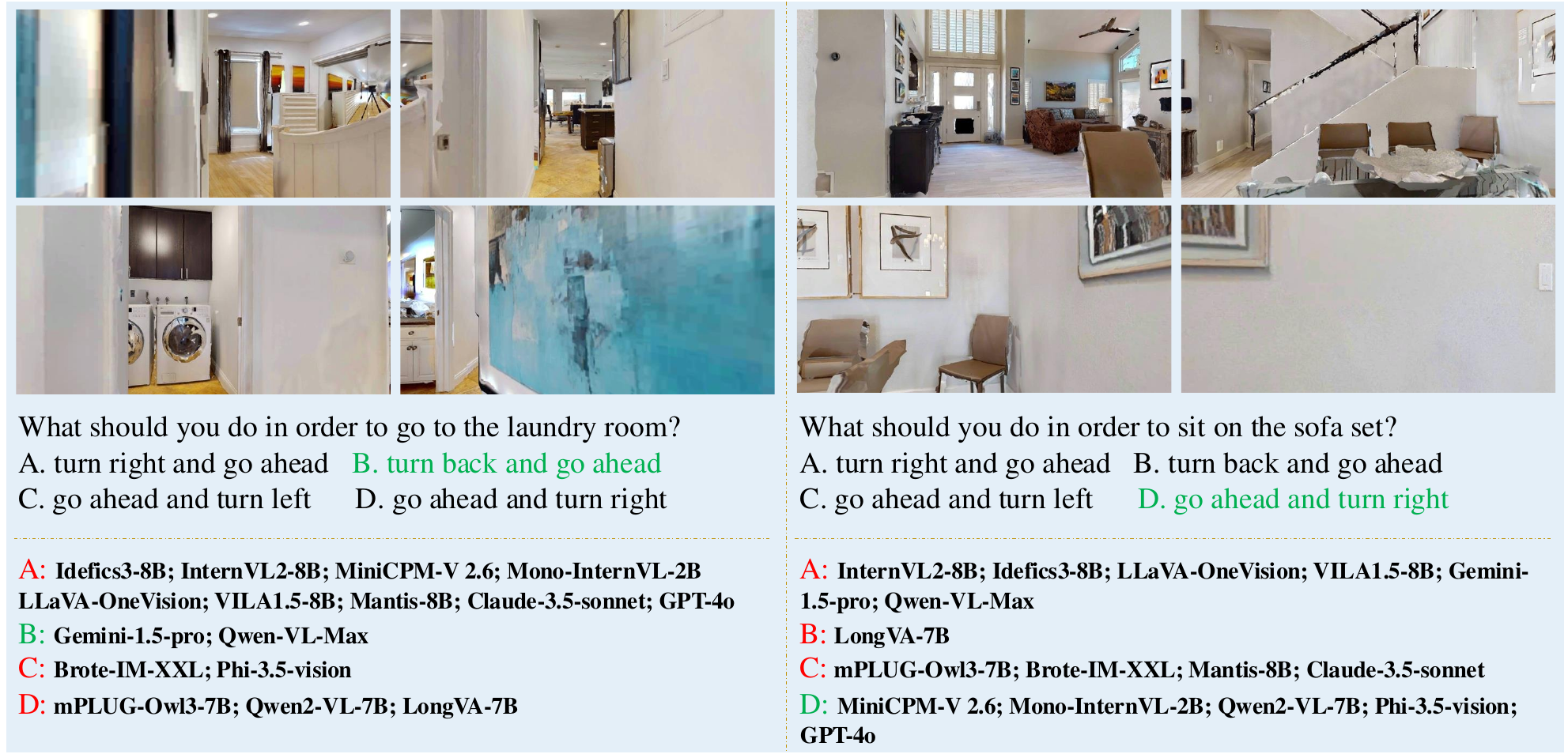}
  \caption{Cases for Planning Decision task. Images are arranged as the following order: the first image is facing towards front, second facing right, third facing back and fourth facing left.}
  \label{fig:case6}
\end{figure*}

 In this section, we provide examples of cases generated by different models on our proposed CoSpace through Figure~\ref{fig:case1} to Figure~\ref{fig:case6}. As shown in Figure~\ref{fig:case5}, for the left case, three of the four assessed proprietary models selected \textit{``B. backleft''} as the answer. In order to correctly answer this question, we should notice that the dining table appears in the third image, which is facing back, and therefore choose from \textit{``B''} and \textit{``D''}. Actually, these models successfully recognize the appearance of the dining table in the third image, but as is located on the left side of the image, they consequently identify the answer as \textit{``B. backleft''}. However, the left side of the third image represents the \textit{``backright''} relative to the standing position in the real space. These models were deceived by the raw visual clues and failed to fill the gap between given images and the original continuous space.

\end{document}